\documentclass[final,5p,times,twocolumn]{elsarticle}

\usepackage{graphicx}
\usepackage{amssymb}

\usepackage{lineno}

\usepackage{amssymb}

\usepackage{amssymb}
\usepackage{amsmath}
\usepackage{siunitx}
\usepackage{tabularx}
\usepackage{algorithmic}

\usepackage{algorithm2e} 
\usepackage{amsthm}
\usepackage{graphicx}
\usepackage{lscape}
\usepackage{verbatim}
\usepackage{color,soul}
\usepackage{xcolor}
\usepackage{subfigure}
\usepackage{tabularx,ragged2e,booktabs,caption,array,multirow,multicol}
\usepackage{csquotes}
\usepackage{mathtools}
\usepackage{lineno}
\usepackage{amsthm}  
\usepackage{listings}
\usepackage{amssymb}
\usepackage{latexsym}
\usepackage{epsfig}
\usepackage{float}
\usepackage{xspace} 
\usepackage{subfig}
\usepackage{url}

\journal{Environmental Modelling \& Software}

\begin{document}

\begin{frontmatter}


\title{ Delhi air quality prediction using LSTM deep learning models with a focus on COVID-19 lockdown}

\author[iitg]{Animesh Tiwari}
\ead{animesh18a@iitg.ac.in}

\author[iitk]{Rishabh Gupta}
\ead{rishabhgupta05@gmail.com}

\author[unsw]{Rohitash Chandra}
\ead{rohitash.chandra@sydney.edu.au}

\address[iitg]{Department of Civil Engineering, Indian Institute of Technology Guwahati, Guwahati, India}

\address[iitk]{Department of Geology and Geophysics, Indian Institute of Technology Kharagpur, Kharagpur, India}

\address[unsw]{School of Mathematics and Statistics, UNSW  Sydney, 
NSW 2006, Australia}

\begin{abstract}
  Air pollution has a wide range of implications on  agriculture, economy, road accidents, and health.   In this paper, we use novel deep learning  methods for short-term (multi-step-ahead) air-quality prediction in selected parts of Delhi, India. Our deep learning methods comprise  of  long short-term memory (LSTM) network models which also include some recent versions such as  bidirectional-LSTM  and   encoder-decoder LSTM models.  We use a multivariate time series approach that attempts to predict air quality for 10 prediction horizons covering total of 80 hours and provide a long-term (one month ahead) forecast with uncertainties quantified.    Our results show that the multivariate bidirectional-LSTM model provides best predictions despite COVID-19 impact on the air-quality  during full and partial lockdown periods.   The effect of COVID-19   on the air quality has been significant  during  full lockdown; however,  there was unprecedented growth of poor air quality afterwards. 
\end{abstract}


\end{frontmatter}


\section{Introduction}
\label{S:1}

The global human population has risen by more than four times during the last century \cite{MaxRoser2013}. Research show that the major  growth in population is attributed to the metropolitan areas in the less developed regions around the globe \cite{desa2015united,taylor2004world}. The consequence of these increased levels  of growth in poorly developed states  is low air quality. We note that 80\% of global cities \cite{taylor2004world} and 98\% of cities in middle-income countries surpass the proposed levels of air quality \cite{who2016,unep2016}.  Increase in air pollution results in economic losses, reduced visibility, contributes to faster climate change that contribute to extreme weather conditions, millions of premature deaths annually \cite{harlan2011climate}. The major factor in air pollution is the anthropogenic fine particulate matter (PM); i.e. PM2.5 (particles with an aerodynamic diameter shorter than 2.5 micrometer)  \cite{gupta2006satellite,zhang2015fine,liu2015transparent,franklin2007association}.  Despite the concentrations of PM2.5 being two to five times higher in developing countries, most of the air quality considerations and estimations are analyzed for developed countries
 \cite{lelieveld2015contribution}.

 Delhi is one of the most prominent  in terms of  growing cities has an estimated   population of more than 19.3 million     \cite{indiapop}. The population density and growth in the last few decades  and rapid industrial expansion  led to massive air pollution to hazardous levels and thus failed in providing people with one of the primary life amenities, quality  air. The \textit{World Economic Forum} recently reported  India having  6 of the world’s 10 most polluted cities with Delhi has one of most polluted \cite{weforum2020}. Research shows  that infected air is one of the prominent causes of premature deaths \cite{who2020} and  the average life span reduces  due to increasing rates of air contamination \cite{apte2018ambient}. It is well known that apart from industrial pollution, agricultural fires are one of the significant contributors of air pollution in Delhi \cite{cusworth2018quantifying}.


Apart from strategies that focus on management of industrial waste\cite{hogland2000assessment}, it is important to model and forecast the air-quality both in short and long-term. Machine learning methods have been promoting for forecasting temporal sequences and  their application to air-quality forecasting has gained  attention recently \cite{
le2020spatiotemporal,li2016deep,bui2018deep}. Forecasting models can be used to develop strategies to evaluate and alarm the general public for future hazardous levels of air quality index. 
Forecasting models for  air pollution concentrations can be broadly classified into two major categories; simulation based and data driven approaches such as statistical or machine learning methods. Simulation-based
method incorporates physical  and chemical models  for generating meteorological
and background parameters to
simulate emission, transport and chemical transformation
of air pollution \cite{grell2005,emmons2009}].
However, they suffer from numerical model
uncertainties and due to the lack of data, the parameterization
of aerosol emissions is restricted \cite{karimian2016hamed}].
Data driven approaches exploit statistical and machine
learning techniques to detect patterns between predictors
and dependent variables in temporal sequences \cite{Yuan2017PredictingTA,10.1145/1390156.1390177, fan2008,inproceedings,lindbeck2000}. Machine learning methods can be used to identify the exposures relevant to health outcomes of interest within high-dimensional  data \cite{patel2017analytic}. Advances in deep learning methods give further motivations for application to domain of air-quality prediction. 

Recurrent neural networks (RNNs) are prominent deep learning models \cite{schmidhuber2015deep}  suited for  modelling temporal sequences, especially those involving long-term dependencies \cite{elman_Zipser1988,Werbos_1990,hochreiter1997long,schmidhuber2015deep,ChandraTNNLS2015}.   Long Short Term Memory networks (LSTMs) were developed \cite{hochreiter1997long} to address limitations in   learning long-term dependencies in sequences   by canonical RNNs \cite{hochreiter1998vanishing,bengio1994learning}.   Gated Recurrent Unit (GRU) \cite{chung2014empirical,cho2014learning} networks are more simpler to implement but provides similar performance than LSTMs. Bi-directional RNNs connect two hidden layers of opposite directions to the same output where the output layer can get information from past and future states simultaneously \cite{650093}.  This led to  bidirectional-LSTMs  for  phoneme classification \cite{10.1016/j.neunet.2005.06.042} which performed better than standard RNNs and  LSTMs and has the potential to be used for forecasting air-pollution time series.

The \textit{coronavirus disease 2019} (COVID-19) is an infectious disease    \cite{Gorbalenya2020species,monteil2020inhibition,world2020coronavirus}    which became a global pandemic \cite{cucinotta2020declares} with major impact to our lifestyle. COVID-19   forced many countries  to close their borders and enforce a partial or full lock down with  devastating impact on the world economy   \cite{atkeson2020will,fernandes2020economic,maliszewska2020potential}. There has been studies relating the effect of environment and air pollution on COVID-19 and vice versa \cite{kerimray2020assessing,dantas2020impact,dantas2020impact} with studies regarding China  \cite{yongjian2020association}, Kazakhstan \cite{kerimray2020assessing} and \cite{dantas2020impact} Brazil. In some cases, COVID-19 lock-downs with reduced traffic showed to reduce air pollution   while in others, it did not make significant impact due to meteorological conditions and factors such as industrial pollutants  \cite{wang2020severe,dantas2020impact}.  Although machine learning has been used for forecasting air-quality, there is scope in improving the forecasts in using latest machine learning models, that feature deep learning models such as recurrent neural networks. Given this  motivation, we focus on air-quality index of Delhi which lately reached hazardous levels. We note that the air-quality in Delhi has significantly improved post COVID-19 pandemic \cite{bbcDelhi2020}, however this was due to lock-downs by the government and the air-quality can deteriorate further when the restrictions are eased. Hence it is important to develop robust forecasting models, that are applicable even during lock-downs and eras of lock-downs during the COVID-19 pandemic.

 In this paper, use novel deep learning  methods for long-term air-quality prediction in selected parts of Delhi, India. Our deep learning methods comprise  of  Long Short Term Memory (LSTMs) networks, bidirectional-LSTMs  and   encoder-decoder LSTMs.  We use a multivariate time series approach that attempts to predict air quality for 10 prediction horizons covering a total of 80 hours with 8 hours for each prediction horizon. We also provide a long-term (one month ahead) forecast with uncertainty quantification given feedback of predicted values into the model. We  first show visualisation of the air quality indicators before and during  coronavirus (COVID) restrictions in Delhi. We  investigate the impact of COVID-19 on air-quality  and the ability of the model to provide quality predictions before and during COVID-19.   Our   models feature data that considers major seasons,  effect of COVID-19, and considers multivariate and univariate  approach for predictions.

The rest of the paper is organised as follows. Section 2 presents a background and literature review of related work. Section 3 presents the proposed methodology and Section 4 presents experiments and results. Section 5 provides a discussion and Section 6 concludes the paper with discussion of future work.

\section{Related Work}


 \subsection{COVID-19 and air pollution}
  
There have been studies relating the effect of environment and air pollution on COVID-19 and vice versa \cite{kerimray2020assessing,dantas2020impact,dantas2020impact}. Zhu et  al.  \cite{yongjian2020association} found that  there was a significant relationship between air pollution and COVID-19 infection in China. Kerimray et  al. \cite{kerimray2020assessing} presented an assessment  on the impact of COVID-19 in large cities in Kazakhstan and found that the temporal reduction in pollution may not be directly attributed to the lockdown due to favorable meteorological variations during the period, however the spatial effects of the lockdown on the pollution levels were clear. Moreover, other non-traffic based sources such as coal based power plants  substantially contributed  to the pollution level.  Dantas et al. 
\cite{dantas2020impact} presented a study on the impact of COVID-19 partial lockdown on the air quality of the city of
Rio de Janeiro, Brazil. The authors reported that 
 The carbon-dioxide  levels showed the most significant reductions during the partial lockdown while nitrogen-oxide decreased in a lower extent, due to industrial and diesel input. The air quality index (with PM-10 concentration) was only reduced during the first partial lockout week. Bao and Zhang \cite{BAO202013} reviewed 44 cities in China to find if COVID-19 lockdowns had an effect on air pollution.  The authors showed  that the lockdowns of 44 cities reduced human movements by 69.85 \%, and the reduction in the air quality index was mediated by human mobility. 
Li et  al. \cite{li2020air} presented a study on air quality changes during the COVID-19 lockdown over the Yangtze River Delta region in Northern China where the authors showed that the ozone did not show any reduction. Moreover,  even during the lockdown it was evident  that background and residual pollution are still high, which includes sources mostly from the industry.  Dutheil and Naval \cite{dutheil2020covid} investigated  COVID-19 as a factor influencing air pollution  for China. The authors argued that  COVID-19 pandemic might paradoxically have decreased the total number of deaths during the period by drastically decreasing the number of fatalities due to air pollution, apart from positive benefits in reducing preventable non communicable diseases. Wang et al. \cite{wang2020severe} showed that in case of China, severe air pollution events were not avoided by reduced  activities during COVID-19 outbreak due to  adverse meteorological events. The authors  highlight that large emissions reduction in transportation and slight reduction in industrial would not help avoid severe air pollution  especially when meteorology is unfavorable.

There have been studies if there is another correlation with rise in COVID-19  infections to the weather change. Tosepu et al. \cite{tosepu2020correlation} presented a study regarding the correlation between weather and COVID-19 pandemic in Jakarta, Indonesia taking into account different temperature levels and percentage of humidity, and amount of  rainfall.  The authors reported that only the average  temperature  was significantly correlated with COVID-19 pandemic. Ma et  al. \cite{ma2020effects} presented a study on the  effects of temperature variation and humidity on the death of COVID-19 in Wuhan, China. The authors collected the daily death numbers of COVID-19 and meteorological parameters and air pollutant data and used generalized additive model  to explore the effect of temperature, humidity and diurnal temperature range on the daily death counts of COVID-19.  The authors  reported that the temperature variation and humidity may be important factors affecting the COVID-19 mortality.

 \subsection{Machine learning methods for forecasting air pollution}

 We review machine learning methods used for  air
quality forecasting in the past decades.  Machine learning methods  have  achieved tremendous success
in a variety of areas for air quality forecasting \cite{Yuan2017PredictingTA,10.1145/1390156.1390177, fan2008,inproceedings,lindbeck2000}.  Although neural networks have advantages over
traditional statistical methods in air quality forecasting,  they have room for improvements \cite{wang2003} due to challenges that include computational
expense, sub-optimal convergence, over-fitting,  and  noisy data. Moreover, the challenge is in configuration of the network topology and model parameters  which   affects the prediction performance. Corani \cite{corani2005}  used models to predict hourly   PM-10 concentrations on the basis of data from
the previous day with   feedforward  
and pruned neural networks.  Jiang et al. \cite{jiang2004d} explored
multiple models that feature physical and chemical model, regression model, and multiple layer perceptron on
the air pollutant prediction task, and their results show that statistical models are competitive with the
classical physical and chemical models.

Machine learning is an active field of research and every now and then novel tools and techniques  emerge for more refined modeling of a specific problem. Some recent works use straightforward approaches like box models, Gaussian models and linear statistical models, which  are easy to implement
and allow for the rapid calculation of forecasts \cite{Peng2015AirQP}. Fu et al. \cite{fu2015} applied
a rolling mechanism and gray model to improve traditional neural network models. Chang et al. \cite{CHANG20201451} used aggregated LSTM  networks and compared the results with  support vector regression, and gradient boosted tree regression. 
 Karimian et  al. \cite{karimian2016hamed} implemented three machine learning methods to forecast air quality given by PM-2.5 concentrations over
different time intervals; these include multiple additive regression trees,  deep feedforward neural network, and a hybrid model based on  LSTM  networks, where  the LSTM model was most  effective for forecasting and controlling air pollution.  Xiao et  al. \cite{xiao2018} parameterised non-intrusive reduced order model  based on proper orthogonal decomposition   for model reduction of pollutant transport equations  to provide rapid response urban air pollution predictions and controls.  Zhu et  al. \cite{zhu2018}  focused on refined modeling for predicting hourly air pollutant concentrations
on the basis of historical meteorological and air pollution data.


 
 \subsection{Effect of air-pollution}

Data-driven machine learning methods present an opportunity to simultaneously assess the impact of multiple air pollutants on health outcomes. There is growing evidence that early-life exposure to ambient air pollution  affect neuro development in children \cite{kim2014prenatal}.  A study that showed that air pollution is
a potential risk factor for obesity with higher body-mass index  in adults that warrants further investigation about other health effects\cite{HUANG2020}. Assessing schoolchildren's exposure to air pollution during the daily commute in a systematic review highlighted studies with schoolchildren's exposure during commutes that are linked with adverse cognitive outcomes and severe wheeze in asthmatic children\cite{MA2020}. Furthermore, Ambient air pollution   associated with reduction in lung functionality and other respiratory conditions  among children \cite{LIU2020}.

Apart from health, air pollution has taken a toll on various other sectors, which includes agriculture.   Industrial air pollution  has a drastic effect in agricultural production as shown by a study in China with lower  marginal products and further alters the relationships of labour-capital  and other factors \cite{WANG2020}. Finally, air pollution has a drastic effect on development and economy. A study on the relationship between air pollution and stock returns  further showed that  industrial air pollution significantly reduces the technical efficiency of agricultural production \cite{XU2020x}.

\section{Methodology }

\subsection{Delhi Situation Report and Data}

 According to the United Nations 2016 report, Delhi had an estimated population of 26 million in greater metro area which is projected to rise to 36 million by 2030 and will remain second most populated city after Tokyo \cite{UNworldcities2016}. The current population density is also one of the worlds largest that will continue to pose further challenges to air quality and health. According to a study by Indian Ministry of Earth Sciences in 2018, it was shown that number of vehicles increases by four times since 2000 which has been a major factor for air pollution in India which includes PM2.5 and hazardous nitrogen-oxide \cite{indiatimesvehicles2020}. In the past five years, the air quality index of Delhi has been  generally moderate (101–200) level between January to September. The air quality index then drastically deteriorates to very poor (301–400), and further  Severe (401–500) or Hazardous (500+) levels during October to December due to various factors  \cite{delhiJanApril,delhifeb}. The air pollution status in Delhi has undergone many changes in terms of the levels of pollutants and the control measures taken to reduce them. Sulian et  al. \cite{sulian2013} provides an evidence-based insight into the status of air pollution in Delhi and its effects on health and control measures instituted. 
  

The meteorological conditions, such as regional and synoptic meteorology are significant in determining the air pollutant concentrations\cite{ab2ef8b1142c4a86ae007857626db575,article,10.1175/1520-0450(1994)033<1182:AACSDT>2.0.CO;2,1997AtmEn..31..869Z}. Lower wind speed (weak dispersion/ventilation) can result in higher concentrations of traffic pollutants  \cite{watson1988atmospheric}.
 However, strong wind speed might form dust storms and end up blowing the particles on the ground.\cite{natsadorj2003}. 
 Higher humidity levels are associated with higher aggregates of air pollutants like PM2.5, carbon monoxide (CO), nitrogen dioxide (NO2) and sulfur dioxide (SO2)   \cite{zalakeviciute2018contrasted}. 
 In addition, high levels of humidity often indicate precipitation events which result in heavy wet deposition leading to the lowering of air pollutants\cite{seinfeld2016}. 
 The most important factors of attenuated visibility are the interactions of particle compositions and light \cite{1985AtmEn..19.1525A,deng2008}. Thus, low visibility can be considered as a strong indicator of high PM2.5 concentrations. 
 Many a time, the formation of some significant air pollutants like ozone (O3) is reduced by cloud cover as they absorb and scatter solar radiations 
\cite{10.1175/1520-0469(1977)034<1149:TIOPOT>2.0.CO;2,akbari2002}. Therefore,  these meteorological variables are importantly selected to predict air quality and hence we take them into consideration.

We   chose  four most polluted areas in Delhi and several surrounding districts   known as the National Capital Region (NCR)   for our study. Bawana was the most-polluted area with an air quality index  of 497, followed by DTU-Delhi Technological University (487), 
Anand Vihar (484) and Vivek Vihar (482) as shown in Figure \ref{fig:delhi_map}. There are   12 parameters that have been considered for our study, the data is taken from  the Central Pollution Control Board, India \cite{CCR:2020} for 3 years in 8 hour interval. The dataset consists of concentration levels of different parameters of air-pollution from four different monitoring stations installed by the government agency in Delhi-NCR region. Tables [\ref{tab:param_anand}, \ref{tab:param_bawana}, \ref{tab:param_dtu}, \ref{tab:param_vivek}] show the range of values for different parameters in our dataset between 1 January,2019 and 10 December, 2020 for different monitoring stations. We find the monitoring station at Bawana records the highest range of values and mean for PM2.5 concentration as compared to other monitoring stations.

\begin{table}[htbp!]
\small
\begin{tabular}{ p{3.0cm} p{2.0cm} p{2cm}  }
  
 \multicolumn{3}{c}{ } \\
 \hline
 Parameter  & Range(Min-Max) & Mean and std\\
 \hline
 PM10 &14.19-939.30 & 238.95 158.72\\
 Benzene    &0-30.82 & 3.86  3.41\\
 Toluene&   0-450.98  & 32.06 40.87\\
 Ammonia Gas(NH3)& 0.10-136.30  & 44.78 23.78 \\
 Nitric Acid(NO) & 0.3-490.90  & 80.60 78.38\\
 Nitrous Oxide(NO2)  & 0.43-360.51  & 74.19 41.04\\
 Nitrogen Oxides(NOx) & 0.13-472.48  & 107.19 80.08\\
 Wind Speed(WS) & 0.3-4.43  & 0.86 0.68\\
 Ozone & 1.42-180.38  & 36.15 26.34\\
 Sulphur Dioxide(SO2) & 0.50-94.36  & 13.57 9.50\\
 Carbon Monoxide(CO) & 0.01-7.21  & 2.14 1.10\\
 PM2.5 & 6.93-936.33  & 116.83 106.76\\
 \hline
\end{tabular}
\caption{Details about features for Anand Vihar.}
\label{tab:param_anand}
   \end{table}
\begin{table}[htbp!]
\small
\begin{tabular}{ p{3.0cm} p{2.0cm} p{2cm}  }
  
 \multicolumn{3}{c}{ } \\
 \hline
 Parameter  & Range & Mean and std\\
 \hline
 PM10 &6.88-941.00 & 227.86 146.87\\
 Benzene    &0.1-0.43 & 0.26 0.01\\
 Toluene&   1.0-354.83  & 137.58 26.89\\
 Ammonia Gas(NH3)& 0.03-127.2  & 36.45 15.99 \\
 Nitric Acid(NO) & 0.27-180.95  & 10.28 18.56\\
 Nitrous Oxide(NO2)  & 5.37-230.82  & 39.91 25.83\\
 Nitrogen Oxides(NOx) & 4.32-184.36  & 29.62 24.63\\
 Wind Speed(WS) & 0.3-4.11  & 1.18  0.55\\
 Ozone & 0.1-170.25  & 44.81 35.06\\
 Sulphur Dioxide(SO2) & 0.56-58.02  & 12.76 8.29\\
 Carbon Monoxide(CO) & 0.0-6.90  & 1.22 0.80\\
 PM2.5 & 10.73- 994.00  & 123.37 109.30\\
 \hline
\end{tabular}
\caption{Details about features for Bawana.}
\label{tab:param_bawana}
   \end{table}
\begin{table}[htbp!]
\small
\begin{tabular}{ p{3.0cm} p{2.0cm} p{2cm}  }
  
 \multicolumn{3}{c}{ } \\
 \hline
 Parameter  & Range & Mean and std\\
 \hline
 PM10 &13.00-1000.0 & 218.05 138.61\\
 Benzene    &0.03-40.86 & 4.96 4.20\\
 Toluene&   10.37-238.50  & 43.02 9.18\\
 Ammonia Gas(NH3)& 0.18-300.65  & 25.84 20.60 \\
 Nitric Acid(NO) & 0.79-351.21  & 18.50 23.56\\
 Nitrous Oxide(NO2)  & 2.00-265.82  & 38.45 26.99\\
 Nitrogen Oxides(NOx)  & 0.0-344.35  & 34.42 29.82\\
 
 Wind Speed(WS) &0.23-6.24  & 0.69 0.37\\
 Ozone & 1.66-186.68  & 51.60 30.84\\
 Sulphur Dioxide(SO2) & 0.52-78.14  & 10.12 7.81\\
 Carbon Monoxide(CO) & 0.0-7.54  & 0.90\\
 PM2.5 & 9.17-948.29  & 111.29 98.01\\
 \hline
\end{tabular}
\caption{Details about features for DTU.}
\label{tab:param_dtu}
   \end{table}
\begin{table}[htbp!]
\small
\begin{tabular}{ p{3.0cm} p{2.0cm} p{2cm}  }
  
 \multicolumn{3}{c}{ } \\
 \hline
 Parameter  & Range & Mean and std\\
 \hline
 PM10 &50.00-994.00 & 211.11 137.77 \\
 Benzene    &0.12-22.54 & 4.43 3.52\\
 Toluene&   3.17-340.20  & 44.48 42.43\\
 Ammonia Gas(NH3)& 4.40-206.48  & 33.80 22.55 \\
 Nitric Acid(NO) & 0.38-440.74  & 18.19 37.06\\
 Nitrous Oxide(NO2)  & 0.20-130.12  & 33.49 19.25\\
 Nitrogen Oxides(NOx) & 4.29-428.66  & 32.73 35.49\\
 Wind Speed(WS) & 0.20-15.0   & 1.69 1.35\\
 Ozone & 0.31-180.5  & 39.63 29.00\\
 Sulphur Dioxide(SO2) & 2.08-108.51 & 22.14 11.63\\
 Carbon Monoxide(CO) & 0.0-7.84  & 1.56 0.97\\
 PM2.5 & 3.54 907.00  & 109.69 104.16\\
 \hline
\end{tabular}
\caption{Details about features for Vivek Vihar.}
\label{tab:param_vivek}
   \end{table}
  \begin{figure*}[htbp!]
  \begin{center}  
   \includegraphics[width=180mm]{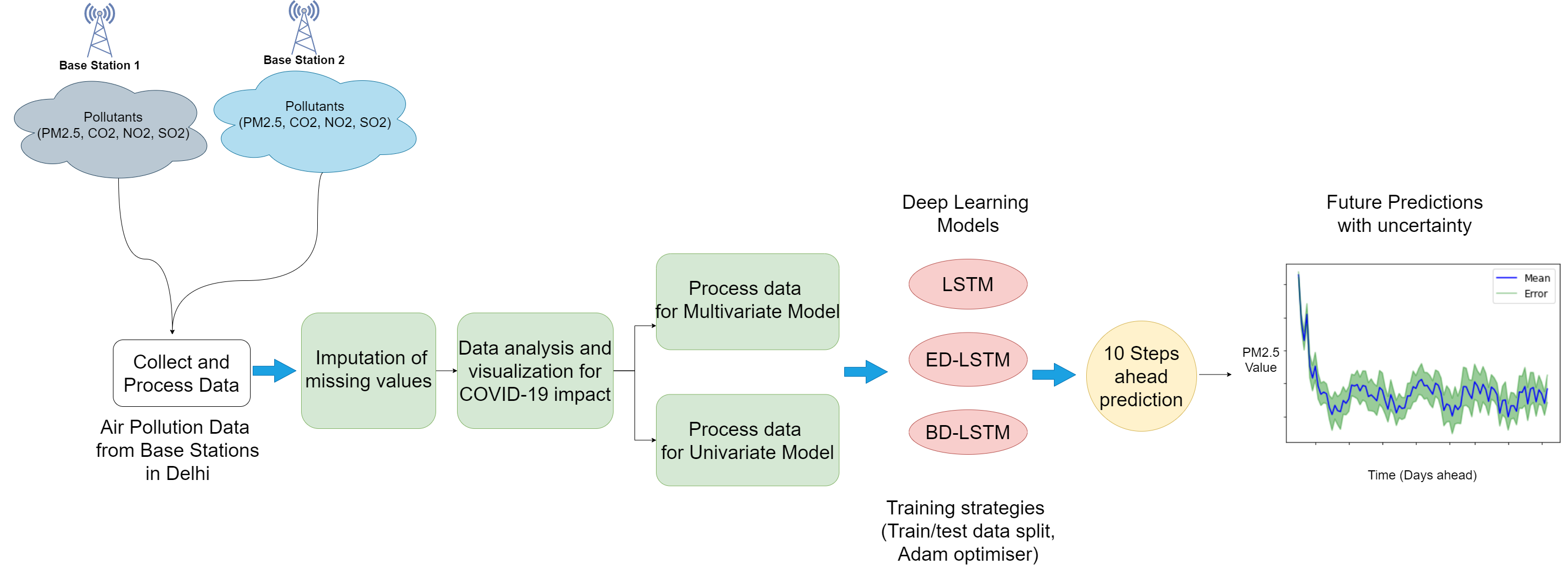} \\
    \caption{Framework diagram for selected monitoring base stations and selected pollutants showing the data collection from base stations, pre-processing and feeding of pre-processed data into the deep learning models for predicting PM2.5 values with uncertainty. We use three deep learning models that include, LSTM, BD-LSTM and ED-LSTM for multivariate and univariate models as shown. }
 \label{fig:framework}
  \end{center}
\end{figure*}
 \subsection{LSTM network models}
  
 Recurrent Neural Networks (RNNs) are a class of artificial neural networks which gained a lot of popularity in recent years. The Elman RNN \cite{elman_Zipser1988,Elman_1990} is one of the earliest architectures and considered a prominent example of \textit{simple recurrent networks}  trained by back-propagation through-time (BPTT) algorithm  \cite{Werbos_1990}. In the early days, there was much research for modeling dynamic systems using simple RNNs and it was shown that RNNs perform better than feed-forward networks in knowledge representation tasks    \cite{Omlin_thonberetal1996, Omlin_Giles1992,Giles_etal1999}. 
   
 Backpropagation through time (BPTT) \cite{Werbos_1990}    features  error backpropagation which uses the idea of training RNNs using gradient descent in a way similar to feedforward neural networks. The major difference is that the error is back-propagated for a deeper network architecture that features states defined by time; however, BPTT experiences the problems of vanishing and exploding gradients in case of learning tasks which involve long-term dependencies or deep network architectures.  \cite{Hochreiter_1998}.  Long Short Term Memory(LSTM) network \cite{hochreiter1997long}  was developed which was capable of  overcoming the fundamental problems of deep learning  \cite{schmidhuber2015deep}. LSTM networks   addressed the issue with much better capabilities in remembering the   long-term dependencies    using memory cells and gates for temporal sequences. The memory cells are trained in a supervised fashion   using an adaptation of the BPTT algorithm that considers the respective gates \cite{hochreiter1997long}. More recently, Adam optimiser which features adaptive learning rate  has been prominent in training LSTM models via BPTT \cite{kingma2014adam}.

 \subsection{Bi-directional LSTMs}
 
One shortcoming of conventional RNNs is that they are only able to make use of previous context in a sequence to predict future states. Bidirectional
RNNs  \cite{schuster1997} do this by processing the data in both directions with two separate hidden layers, which are then
feed forwards to the same output layer. Bidirectional RNNs are simply two independent RNNs together in a structure that allows the networks to have both backward and forward information about the sequence at every time step.

The combination of  bidirectional RNNs with LSTM model gives us bidirectional LSTM model (BD-LSTM), which have the ability to access long-range context in both input directions \cite{graves2005}.  BD-LSTM are designed for specific input sequences whose starting and ending are known beforehand. They take both the future and past states of each element of a sequence into consideration where one LSTM processes the information from start to end of the sequence and the other from end to start. Their combined outputs predict the corresponding labels if available at each time step \cite{schmidhuber2015deep}. 
This approach differs from unidirectional approach in a way where  information from the future   and past LSTM model states combined; hence, it preserves information from both past and future at each time step. BD-LSTM have been prominent in sequence processing problems such as phoneme classification
\cite{graves2005}, continuous speech recognition \cite{Fan2014TTSSW}, speech synthesis \cite{graves2013hybrid} and sentence classification \cite{ding2018densely}. 
  
  \begin{figure*}[htbp!]
  \begin{center}  
   \includegraphics[width=180mm]{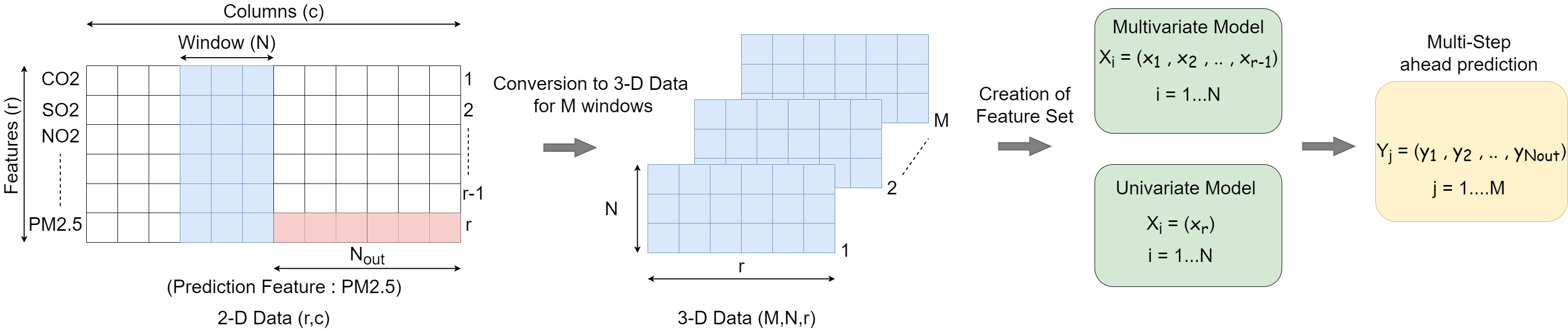} \\
    \caption{Diagram showing the conversion of 2-Dimensional(2-D) data to 3-Dimensional(3-D) data using $M$ windows of size $N$ and creation of feature set for both univariate and multivariate model to give multi-step ahead predictions. For 2-D data, the window is highlighted in blue while the future prediction feature (PM2.5) is highlighted in pink.}
 \label{fig:2D-3D}
  \end{center}
\end{figure*}

 \subsection{Encoder-Decoder LSTMs}
 
  A sequence to sequence model lies behind numerous systems which we face on a daily basis  \cite{NIPS2014_5346}, and such  models aim  to map a fixed-length input with a fixed-length output, where the length of the input and output may differ. In multi-step and multivariate analysis, both the input and output are variable with potentially different lengths. This problem is analogous to machine translation between natural languages, where a sequence of words in the input language is translated to a sequence of words in the output language. Recently,  it has been shown how to effectively address  sequence to sequence problem with encoder-decoder LSTM networks (ED-LSTM). 
 
 ED-LSTM handles variable length inputs and outputs by  first encoding a given input sequence $X_{ip} = (x_1, x_2, .. x_n)$, one at a time using a latent vector representation for input of length $n$ and the output is a sequence $Y = (y_1, y_2, .. y_m)$ of length $m$. In the encoding stage, ED-LSTM creates a sequence of hidden states for input sequence and then in decoding stage it defines a distribution for output states given the input sequence. The main goal of the model is to maximize the conditional probability $P(Y|X_{ip})$ of mapping input sequence to output sequence while training. ED-LSTM have been previously used in a lot of tasks, some of which include text simplification \cite{wang2016experimental}, automatic speech recognition \cite{zeyer2019comparison} and grapheme to phoneme conversion \cite{yao2015sequence}.

\begin{figure}[htbp!]
  \begin{center}  
   \includegraphics[scale=0.33]{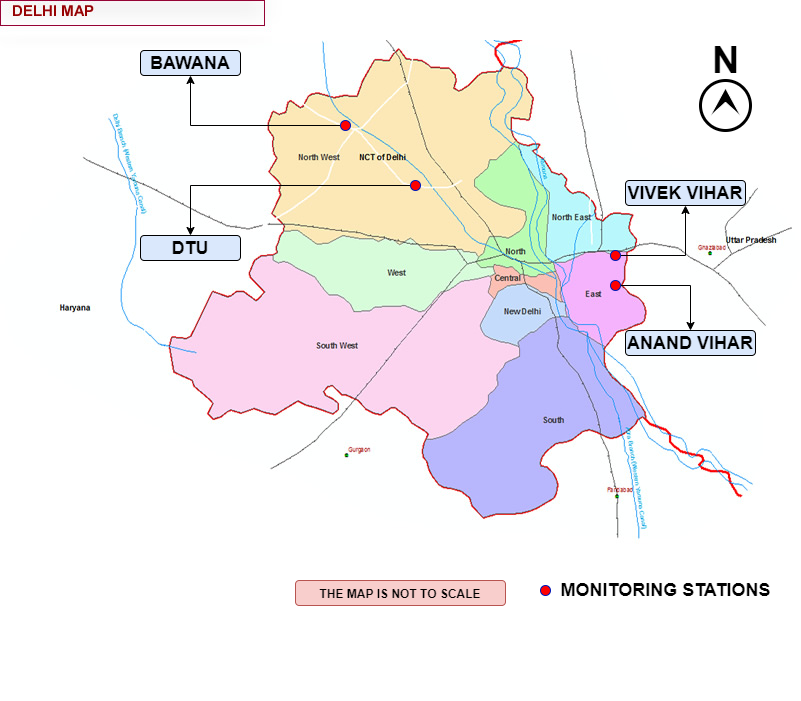} 
    \caption{ Map of Delhi highlighting the location of four monitoring base stations in Delhi.  (Note: The map is not to scale. )}
 \label{fig:delhi_map}
  \end{center}
\end{figure}

\subsection{Framework}
 
The air-pollution dataset for four different monitoring stations has been collected from Central Pollution Control Board (CPCB) \cite{CCR:2020}, Government of India \footnote{\url{https://cpcb.nic.in/}}. Figure \ref{fig:framework} shows the framework diagram for our experimental setup. We pre-process the dataset extracted from four different monitoring stations in Delhi (Figure \ref{fig:delhi_map}   with map adopted   from  website \footnote{\url{http://gismaps.in/prod/indiabasemaps/Delhi/Delhi.html}}) for  our proposed deep learning framework as shown in Figure  \ref{fig:framework}. After collecting the data, we fill the missing values of a particular feature by the median of its available neighboring values. The missing values may be due to inconsistency in recording the pollutant values at different monitoring stations.  In order to train LSTM models, we transform  the two-dimensional (2D) dataset of features and temporal sequences  given by time-interval (eight hours) into a into a three-dimensional  (3D)  format   \cite{masum2018}  by a sliding window technique in which a window of size \textit{N} moves over the original 2D dataset   of size \textit{(r,c)}; where \textit{r} represents the number of rows(features) and \textit{c} represents the number of columns in dataframe. This results in \textit{M} windows  final dataset with dimensions \textit{(M,N,r)} which is fed into the models to give $N_{out}$ multi-step ahead prediction values as shown in Figure \ref{fig:2D-3D}. We break the dataset into a feature set $X_i=({x_1,x_2,x_3....x_{r-1})}, i = 1...N$ for multivariate model and $X_i=({x_r})$ for univariate model and labels $Y_j=({y_1,y_2,...,y_{N_{out}}}), j = 1...M$ as multi-step ahead predictions. The label value consists of PM 2.5 value and other meteorological values constitutes the feature vector for multivariate model while for univariate model past PM2.5 values constitute the feature vector .

 Our framework for the respective LSTM  models is as follows. We use $N_f = 11$ features (different pollutant values other than PM2.5 as shown in Table 1) for our multivariate model where we  consider $N = 5$ a window that captures the  number of time steps in the past. The window   determines how much the LSTM model will unfold in time and the number of features determines the number of input neurons in the respective LSTM models. We define $N_{out}=10$ as the number of output neurons in the LSTM model which denotes   the number of steps ahead  to predict   PM2.5 concentration. We note that each step ahead is denoted by 8 hour intervals and hence the respective models would predict 80 hours ahead. Hence, we use 40 hours (5 steps) of past information to predict 80 hours (10 steps) of future trend of the time series. In the case of univariate model, we only use PM2.5 as a feature to predict its future trend and hence, the respective LSTM models will feature 1 input neuron and 10 output neurons.
  
\begin{table*}[htbp!]
 \small 
 \centering
\begin{tabular}{llllp{8cm}}
\hline
 &  Input Dimension& Hidden Layers & Output & Comments  \\
\hline
\hline
FNN  &      $N\times N_{f} = 55$&2&10& Hidden Layer sizes $(h_1 , h_2)$= (64,32)\\

LSTM & $(N,N_{f}) = (5,11)$ &1&10& Hidden Layer of 50 cells.\\
BD-LSTM & $(N,N_{f}) = (5,11)$&1&10&Forward and Backward layer of 50 Cells each.\\
ED-LSTM& $(N,N_{f}) = (5,11)$&4&10& Two LSTM layers of 50 cells each, Repeat Vector and a Time distributed layer\\

\hline &
\end{tabular}
\caption{Topology of different deep learning models. }
\label{tab:config}
\end{table*}
   
 \section{Experiments and Results}
 In our experiments, we compare the performance of  three LSTM models as presented   earlier for univariate and multivariate prediction models. We first analyse the impact of COVID-19 lock-downs on the air quality and then evaluate the  different models for  multi-step ahead prediction of air quality (PM2.5). Our experiments consider different data setup that features data before and after COVID-19 in order to evaluate the effect  on air quality and model accuracy.

\subsection{Experiment setup}
 
The visualisations and experiments  are organised  as follows. 

\begin{itemize}  

\item Provide analysis of concentration of PM2.5 for different monitoring stations and for different time intervals including the COVID-19 lockdown;
\item Compare the performance of the three LSTM network models using Adam optimizer;
\item Create training data with different strategies, that includes and exclude COVID-19 time-span and compare the performance of best LSTM model; 
\item Compare  univariate and multivariate approach using the best LSTM model;

\item Provide one month forecast using best LSTM model with uncertainty quantification with multiple experimental runs.

\end{itemize}

 The technical details of our framework for the respective LSTM  models as follows. We considered different number of hidden neurons  and other hyper-parameters (such as learning rate) in our trial experiments. We define the topology of different models in terms of input, hidden layers and output in Table \ref{tab:config}. We use  Adam optimizer \cite{kingma2014adam} with batch size of 20 for 200 epochs and rectifier linear unit  (ReLu) activation in the hidden layers for the three LSTM models. We also compare our results with FNN which uses Adam optimiser. 
  
 In the respective experiments, we compare the performance of different models for different time steps by using the train data from Jan 2019 to May 2020 (pre-COVID-19 lockdown) and the test data from June 2020 to Dec 2020 (during and partial COVID-19 lockdown). We test our models by training without shuffling the time-series data window and also by shuffling it.  Shuffling the data here refers to randomly picking different training data windows (regions) comprising of past time steps and its corresponding output time steps. Hence, with shuffling, the training data would feature peak and off-peak regions of COVID-19 as different weeks can be featured in training data rather than taking consecutive weeks.  
 
 Furthermore, we test the best   model in two different ways, first by training it on a seasonal data comprising of observations from February to September in 2019 which excludes the COVID-19 lockdown. We test the  model  on respective months for 2020 which corresponds to the COVID-19 lockdown and partial lockdown periods, in this case we use 50 epoch of training with our best LSTM model. Second, we use a univariate time-series approach for multi-step ahead prediction of PM2.5 values and test the performance using our best model. In this case,  we use 1000 epochs of training with our best LSTM model. In all our experimental setup using different LSTM models and FNN, we further split the test dataset into validation dataset for tuning the model during trial experiments and test our model on the remaining portion for evaluating the models, we split the test dataset in the ratio 1:1 for for all our respective experiments.



The prediction performance is  measured by root mean squared error (RMSE) as follows

$$RMSE = \sqrt{\frac{1}{N} \sum_{i=1}^{N} (y_i - \hat{y_i})^2}$$

\noindent where, $y_i$ and $\hat{y_i}$ are the actual value and the predicted value, respectively. $N$ refers to the total length of the data. We report the RMSE for different horizons and also the mean RMSE for all prediction horizons. 

We conduct 30 independent experimental runs with different weight initialisation in the respective LSTM models and provide mean and standard deviation in all our results.

\begin{figure*}[h!]
\centering
\subfigure[Feature correlation heatmap for Anand Vihar.]{
\includegraphics[scale =0.40]{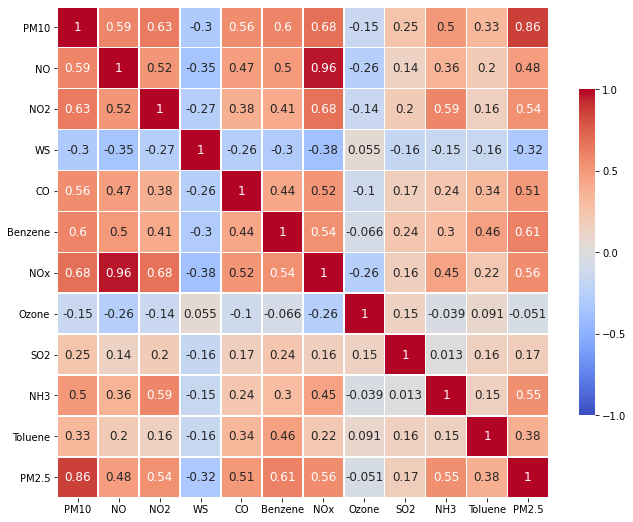}
 }
 \subfigure[Feature correlation heatmap for Bawana.]{
   \includegraphics[scale =0.40] {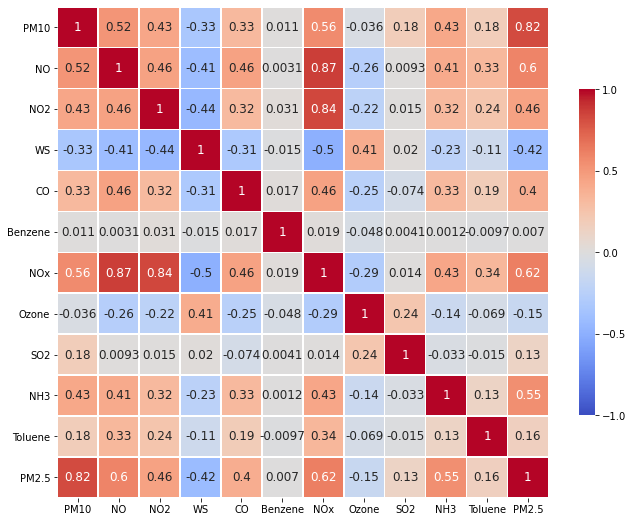}
 }
 \subfigure[Feature correlation heatmap for DTU.]{
   \includegraphics[scale =0.40] {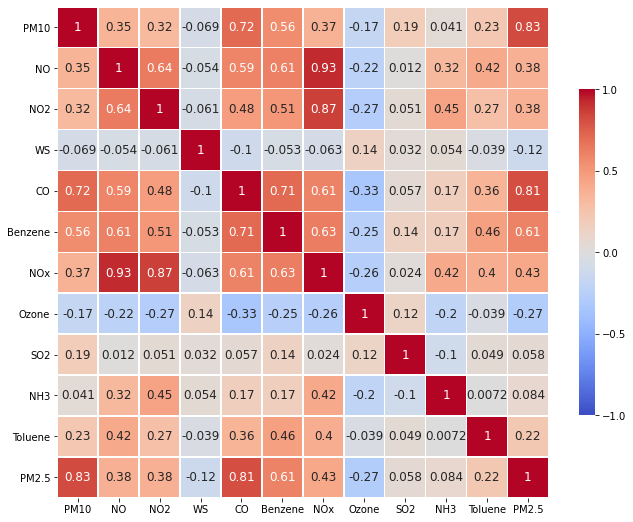}
 }
 \subfigure[Feature correlation heatmap for Vivek Vihar.]{
   \includegraphics[scale =0.40] {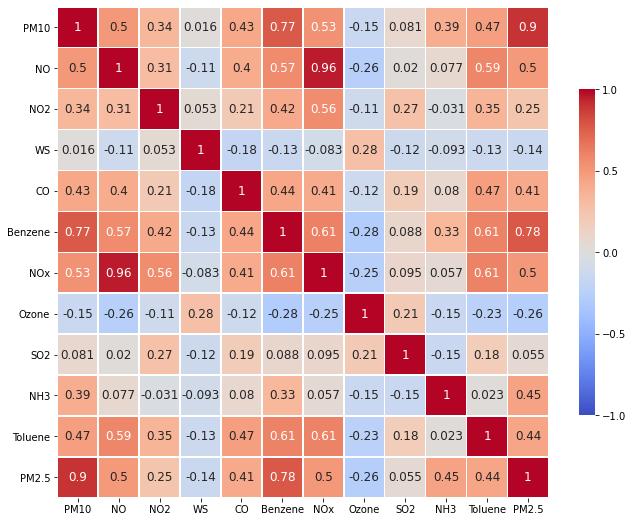}
 }

\caption{Correlation via heat-map  for different features in the data collected from different monitoring stations in Delhi.}

\label{fig:heatmaps}
\end{figure*}

\subsection{ Data visualisation}
 
\begin{figure*}[htbp!]
  \begin{center}  
   \includegraphics[width=190mm]{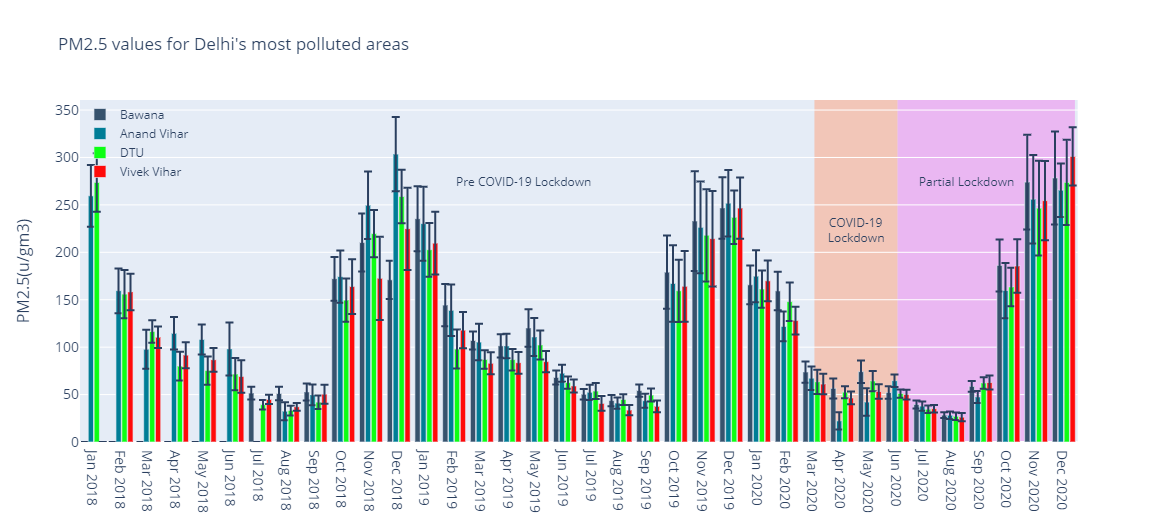} \\
    \caption{ PM2.5 values of Anand Vihar, Bawana, Delhi Technical University, and Wazirpur  from 1 Jan 2018 to 10 Dec 2020. }
 \label{fig:pm2.5}
  \end{center}
\end{figure*}

We first provide data visualisation and analyse the effect of COVID-19 on air pollution in Delhi while comparing with previous years. 
Figure  \ref{fig:pm2.5}  presents PM2.5 values of stations in Anand Vihar, Bawana, Delhi Technical University and Vivek Vihar from 1 January 2018 to 10 December 2020 that is used as dataset for this study. We find some missing data in the initial months of 2018 for some stations therefore we consider the data from 1 January 2019 to 10 December 2020 for building our respective models and strategies.   Table \ref{tab:sa-time-series} presents a summary of the visualisation in Figure \ref{fig:pm2.5}  where the seasons are quantitatively highlighted. We observe higher values of PM2.5 value for the months from October till February on a seasonal basis after which the value starts to descend starting from the month of March. Although we find lower values of PM2.5 during March-June every year,  we observe a significant decrease for   months of interest  (March-June 2020), which belongs to the COVID-19 full lockdown period in India when compared to the respective months in 2018 and 2019as shown in Table \ref{tab:sa-time-series} and Figure \ref{fig:pm2.5}. We find a similar decrease of PM2.5 values for all the monitoring stations while Bawana records the highest values for 2019 and 2020 among other stations. The partial lockdown period include the months after June, 2020  as highlighted in Figure \ref{fig:pm2.5}.

Figure \ref{fig:heatmaps} shows Pearson's correlation heatmaps for different features in our dataset for different monitoring stations. Higher positive values of Pearson's correlation between two particular feature indicates a high positive dependence between them while a higher negative value indicates a high negative dependence between them and the values closer to zero indicates more independent feature sets. Higher positive correlation values are represented as dark brown panels while higher negative correlation values are represented as dark blue panels in different correlation heatmaps. We observe higher correlation values of NOx with NO and NO2 also which is expected because they are nitrogenous pollutants and show higher positive dependence. We also observe high positive correlation values between PM2.5 and PM10 for all the monitoring stations which also indicates a good amount of correlation. 

\begin{table*}[h]
\centering
\smaller
 \caption{Mean with +/- 95\% confidence interval for PM2.5 concentrations for March-June for three consecutive years over different monitoring stations.}
\label{tab:sa-time-series}
\begin{tabular}{llllll}
\hline
 \hline
Time Interval&PM2.5(Bawana)&PM2.5(Anand Vihar)&PM2.5(Vivek Vihar)&PM2.5(DTU)&\\

 &[mean, interval] & [mean, interval]& [mean, interval] & [mean,interval] & \\
\hline
\hline
Mar-Jun(2018)& - - & 104.59 10.47 & 89.55 7.39 & 85.98 8.05  \\
Mar-Jun(2019)& 99.38 7.41 & 97.62 8.48 & 77.65 5.59 & 84.81 6.13  \\
Mar-Jun(2020)& 64.31 5.44 & 49.11 6.45 & 52.78 4.08 & 57.94 4.74  \\
\hline
\hline
\end{tabular}
\end{table*}
 
   
\begin{figure}[htpb!]
\centering
\subfigure[RMSE  for different prediction horizons for Anand Vihar.]{
\includegraphics[scale =0.35]{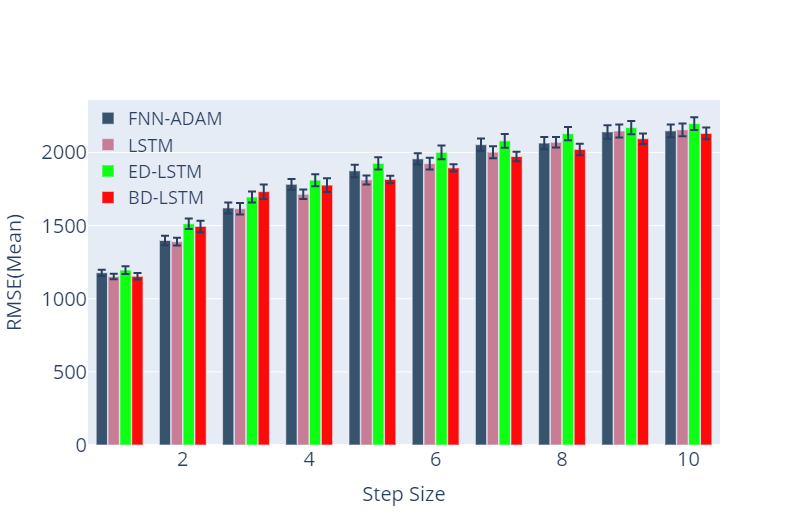}
 }
 \subfigure[RMSE for different  models for Anand Vihar.  ]{
   \includegraphics[scale =0.35] {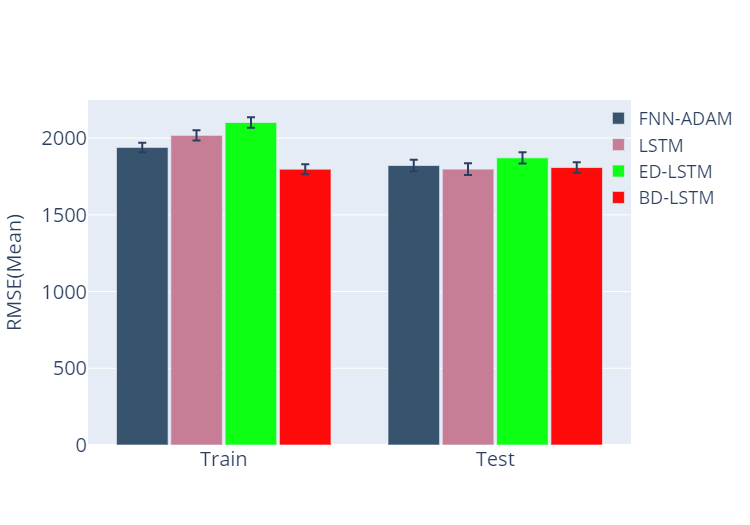}
  
 }
 
\caption{Performance evaluation of respective methods for Anand Vihar (mean RMSE with ($\pm$) 95\% confidence interval  of 30 experimental runs as uncertainty given by error bar).}

\label{fig:trend_models}
\end{figure}

\subsection{Results: Modelling and Forecasting}

 We report the train and test performance (RMSE)   mean and ($\pm$) 95\% interval of the RMSE for different prediction horizons for each model from 30 experimental runs. Table \ref{tab:sa-time-series} shows the mean and ($\pm$) 95\% confidence interval for PM2.5 concentration for different monitoring stations for the months March to June in the years 2018, 2019 and 2020. We selected this particular interval of months to show a comparison of PM2.5 concentration in consecutive years for these COVID-19 lockdown months. We observe a significant decrease in PM2.5 concentration in 2020 for all monitoring stations as compared to 2019 as a consequence of COVID-19 lockdown.     In our first set of experiments, we compare the performance of different  multivariate   LSTM models (LSTM, ED-LSTM and BD-LSTM)  where all the models are trained using Adam Optimizer using the dataset collected from the monitoring station at Anand Vihar. We also show results with canonical feedforward neural network (FNN-Adam) with Adam optimiser. We also show results for  case of shuffling the training data (BD-LSTM*), univariate (UBD-LSTM) and seasonal (SBD-LSTM) at Anand Vihar. Note that all LSTM models are multivariate except for UBD-LSTM.
 
  Figure \ref{fig:trend_models}(a) shows the   prediction horizon RMSE  for 10 step ahead prediction,  and  Figure \ref{fig:trend_models}(b) shows the mean RMSE for entire train and test datasets for Anand Vihar using different LSTM models. We show bar plots with  mean and  ($\pm$) 95\% confidence interval using 30 experimental runs. Note that lower value of RMSE indicates better performance.   We find that the prediction RMSE in general increases across prediction horizons for all the models which is also natural as we use a specified window size as input to predict the multi-step ahead values and the results deteriorate as the gap in the missing values increases with increasing prediction horizon. In Figure \ref{fig:trend_models}(b) and Table \ref{tab:rmse_anand}, we find that the multivariate   BD-LSTM performs the best on train dataset generalizes well with test performance. We find that ED-LSTM is the worst performer in this case with highest RMSE on both train and test dataset in comparison to other models. In Figure \ref{fig:trend_models}(a) and Table \ref{tab:rmse_anand}, we observe that  multivariate  BD-LSTM generally has the best performance which significantly improved for step sizes 6-10 when compared to other models.   

Since all the monitoring stations belong to Delhi region, we use the model which performs the best on dataset from Anand Vihar to analyse the performance on the dataset from other stations namely, Bawana, DTU and Vivek Vihar. 
 Taking into account the results for Anand Vihar, we consider BD-LSTM as our best model,  and further apply it other monitoring stations. We also consider different strategies with BD-LSTM to train and analyse the performance on dataset from different monitoring stations which includes shuffling the input window when training BD-LSTM model (BD-LSTM*), univariate time-series analysis using BD-LSTM model (UBD-LSTM) and training the BD-LSTM model using seasonal data (SBD-LSTM).
 
\begin{figure*}[h!]
\centering
\subfigure[RMSE(mean) with error bar for Anand Vihar.]{
\includegraphics[scale =0.33]{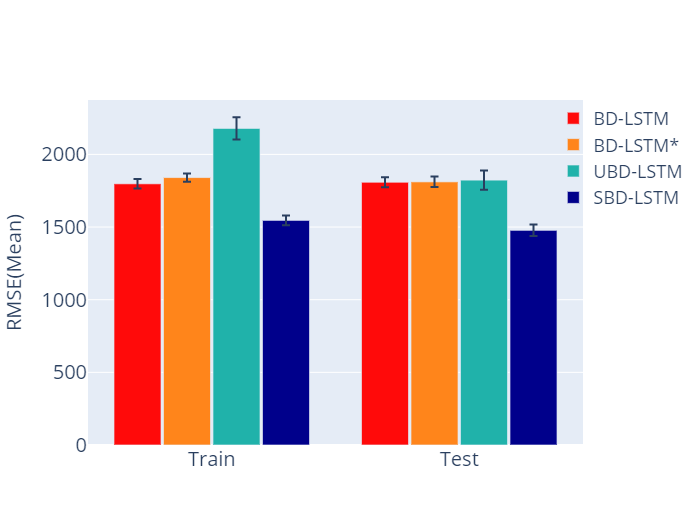}
 }
 \subfigure[RMSE(mean) with error bar for Bawana.]{
   \includegraphics[scale =0.33] {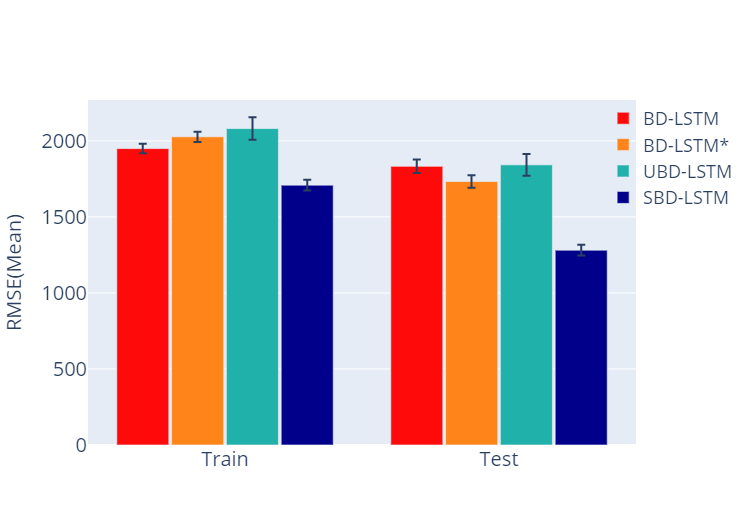}
 }
 \subfigure[RMSE(mean) with error bar for DTU.]{
   \includegraphics[scale =0.33] {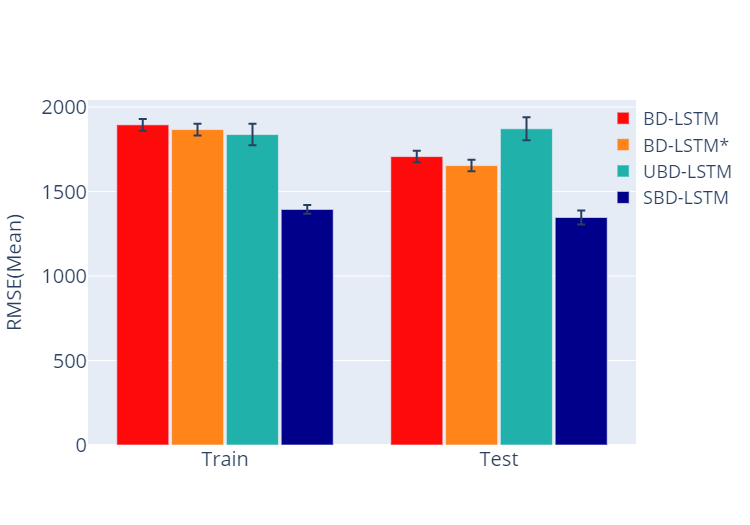}
 }
 \subfigure[RMSE(mean) with error bar for Vivek Vihar.]{
   \includegraphics[scale =0.33] {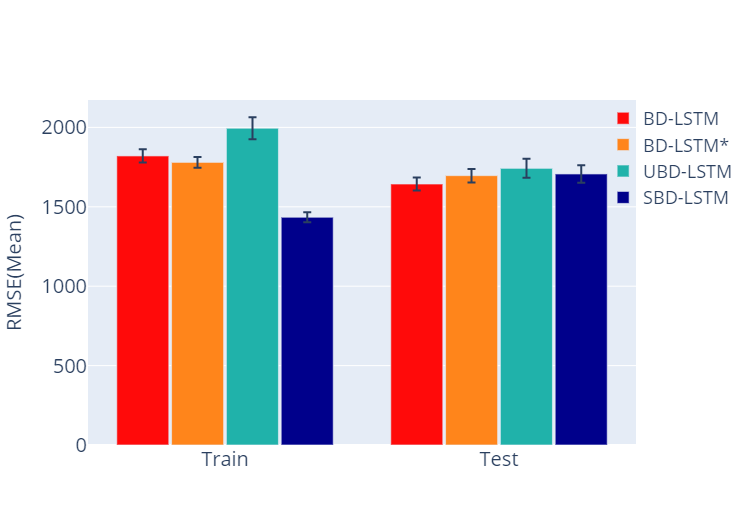}
 }

\caption{Performance by RMSE (mean) with ($\pm$) 95\% confidence interval as error bar across 10 prediction horizons  of 30 experimental runs using respective BD-LSTM models. We show results for train and test datasets collected from different monitoring stations in Delhi.}

\label{fig:lstm_misc}
\end{figure*}

 In Figure \ref{fig:lstm_misc}(a) and Table \ref{tab:rmse_anand}  results for Anand Vihar, we find that BD-LSTM performs better on train dataset in comparison to BD-LSTM* with almost similar performance on test dataset while UBD-LSTM has highest RMSE on both train and test dataset with highest confidence interval(error) of prediction. In Figure \ref{fig:lstm_misc}(b) and Table \ref{tab:rmse_bawana} results for Bawana, we find that BD-LSTM performs the best on train dataset in comparison to BD-LSTM* and UBD-LSTM while BD-LSTM* performs the best on test dataset amongst them. We find a similar trend as Anand Vihar for UBD-LSTM with highest RMSE on both train and test dataset and highest confidence interval of prediction. In Figure \ref{fig:lstm_misc}(c) and Table \ref{tab:rmse_dtu} results for DTU, we find that UBD-LSTM performs better than BD-LSTM and BD-LSTM* on train dataset while BD-LSTM* performs the best on test dataset among these three, we observe a similar trend for UBD-LSTM with highest confidence of prediction on both train and test dataset. In Figure \ref{fig:lstm_misc}(d) and Table \ref{tab:rmse_vivek} results for Vivek Vihar, we find that BD-LSTM* performs the best on train dataset in comparsion to BD-LSTM and UBD-LSTM, while BD-LSTM  performs the best on test dataset among them.  UBD-LSTM has the worst performance on both train and test dataset with largest confidence interval for the predictions.  

 In the  seasonal model  for   the monitoring stations (SBD-LSTM), we find   least RMSE   for both train and test dataset with better performance on test dataset for Anand Vihar, Bawana and DTU. For Vivek Vihar, we find that SBD-LSTM performs the best on train dataset but it does not generalise well  on test dataset. We  next evaluate the performance of SBD-LSTM  in the same plot for other models (BD-LSTM, BD-LSTM*,UBD-LSTM). We cannot make a direct comparison between SBD-LSTM and other models in terms of prediction accuracy, since SBD-LSTM uses a subset of initial data (February-September, 2019) with lower range of PM2.5 values for these months as shown in Figure \ref{fig:pm2.5}.

\begin{figure*}[h!]
\centering
\subfigure[PM2.5 Concentration prediction for next one month for Anand Vihar.]{
\includegraphics[scale =0.60]{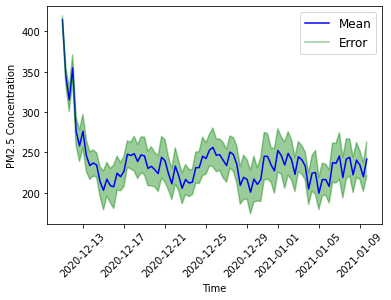}
 }
 \subfigure[PM2.5 Concentration prediction for next one month for Bawana.]{
\includegraphics[scale =0.60]{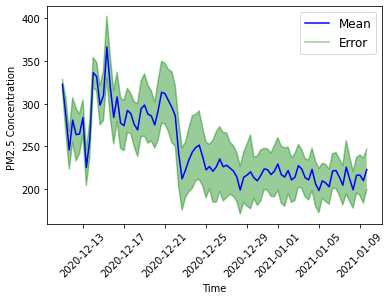}
 }
 \subfigure[PM2.5 Concentration prediction for next one month for DTU.]{
\includegraphics[scale =0.60]{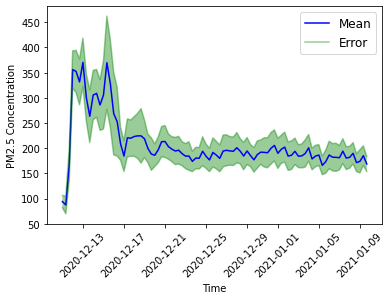}
 }
 \subfigure[PM2.5 Concentration prediction for next one month for Vivek Vihar.]{
\includegraphics[scale =0.60]{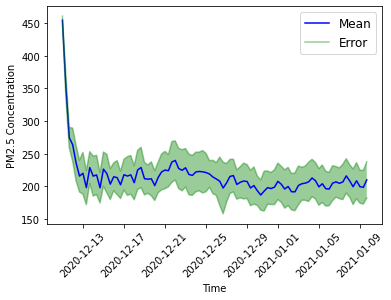}
 }
 
\caption{PM2.5 Concentration prediction with Mean and ($\pm$) 95\% confidence interval as error for next one month using UBD-LSTM model for different monitoring stations.}

\label{fig:future_plots}
\end{figure*}

\begin{figure*}[h!]
\centering
\subfigure[PM2.5 Concentration(actual and predicted) for Anand Vihar.]{
\includegraphics[scale =0.60]{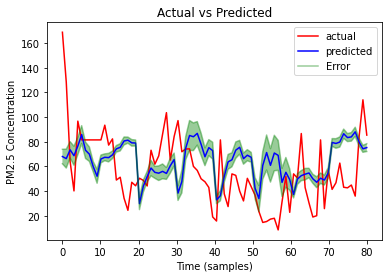}
 }
 \subfigure[PM2.5 Concentration(actual and predicted) for Bawana.]{
\includegraphics[scale =0.60]{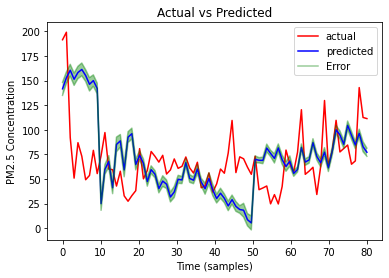}
 }
 \subfigure[PM2.5 Concentration(actual and predicted) for DTU.]{
\includegraphics[scale =0.60]{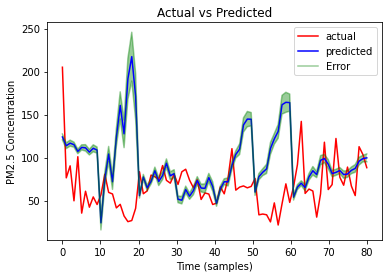}
 }
 \subfigure[PM2.5 Concentration(actual and predicted) for Vivek Vihar.]{
\includegraphics[scale =0.60]{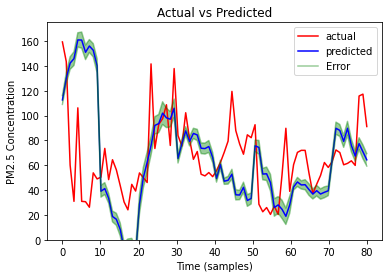}
 }
 
\caption{PM2.5 Concentration (actual and predicted) with Mean and ($\pm$) 95\% confidence interval as error of the predicted values for September, 2020 using SBD-LSTM model for different monitoring stations.}

\label{fig:sept_plots}
\end{figure*}

  Figure \ref{fig:future_plots} provides a long-term (one month  ahead) forecast (11th December 2020 to 9th January 2021) of PM2.5 concentration values for different monitoring stations using UBD-LSTM model. The plots provide mean and ($\pm$) 95\% confidence interval as   uncertainty in predicted values for 30 experimental runs for 8 hours interval for each day, covering a total of 720 hours for the entire  month. Figure \ref{fig:sept_plots} shows a comparison between actual and predicted values of PM2.5 concentration for September 2020 in terms of mean and ($\pm$) 95\% confidence interval  in predicted values for 30 experimental runs for different monitoring stations using SBD-LSTM model.

\begin{table*}[htbp]
 \small 
\begin{tabular}{llllllll}
\hline
 &  FNN-Adam& LSTM & ED-LSTM & BD-LSTM  & BD-LSTM* & UBD-LSTM& SBD-LSTM \\
\hline
\hline
		
Train &      
1939.38$\pm$	30.79&
2017.92$\pm$	33.34&
2101.77$\pm$ 34.16&
1797.43$\pm$ 32.68&
1840.42$\pm$ 28.45&
2178.68$\pm$ 76.78&
1545.79$\pm$ 34.30\\
Test &      
1821.29$\pm$	37.45&
1797.96$\pm$	37.64&
1871.46$\pm$	37.17&
1808.26$\pm$	34.04&
1810.86$\pm$	36.21&
1822.95$\pm$	65.80&
1477.35$\pm$ 39.34\\
Step-1 &      
1177.20$\pm$	21.02&
1152.12$\pm$	19.25&
1195.88$\pm$ 26.43&
1153.02$\pm$ 22.00&
1130.93$\pm$ 19.09&
317.40$\pm$ 28.01&
843.21$\pm$ 27.57\\
Step-2 &      
1397.96$\pm$	32.76&
1390.28$\pm$	26.98&
1512.29$\pm$	36.34&
1494.42$\pm$	39.38&
1504.22$\pm$	50.99&
1451.15$\pm$	58.28&
1034.96$\pm$ 25.71\\
Step-3 &      
1620.34$\pm$	37.52&
1615.66$\pm$	39.30&
1695.36$\pm$	37.67&
1731.97$\pm$	49.72&
1718.83$\pm$	56.19&
1690.91$\pm$	71.69&
1283.02$\pm$ 51.79\\
Step-4 &      
1782.44$\pm$	36.65&
1713.78$\pm$	32.38&
1810.73$\pm$	41.03&
1776.68$\pm$	47.31&
1736.79$\pm$	40.89&
1863.03$\pm$	67.08&
1425.18$\pm$ 68.46\\
Step-5 &      
1873.96$\pm$	42.76&
1812.06$\pm$	31.15&
1925.16$\pm$	41.85&
1815.68$\pm$	25.61&
1806.75$\pm$	46.08&
2010.27$\pm$	62.85&
1508.44$\pm$ 67.65\\
Step-6 &      
1956.12$\pm$	38.44&
1923.46$\pm$	39.53&
2000.59$\pm$	46.58&
1894.14$\pm$	25.17&
1894.19$\pm$	42.10&
2205.94$\pm$	63.72&
1675.28$\pm$ 73.59\\
Step-7 &      
2053.13$\pm$	41.38&
2001.73$\pm$	41.10&
2079.05$\pm$	46.45&
1971.98$\pm$	32.10&
1983.71$\pm$	43.13&
2223.07$\pm$	65.43&
1685.23$\pm$ 56.25\\
Step 8 &      
2064.14$\pm$	41.96&
2069.54$\pm$	36.31&
2128.60$\pm$	45.86&
2020.71$\pm$	39.29&
2054.23$\pm$	45.53&
2174.59$\pm$	44.03&
1718.71$\pm$ 46.06\\
Step 9 &      
2140.37$\pm$	46.04&
2146.61$\pm$	45.23&
2170.13$\pm$	45.64&
2093.84$\pm$	35.45&
2149.13$\pm$	37.42&
2174.37$\pm$	59.44&
1827.85$\pm$ 48.45\\
Step 10 &      
2147.23$\pm$	43.46&
2154.41$\pm$	44.34&
2196.85$\pm$	43.78&
2130.20$\pm$	40.49&
2129.89$\pm$	32.62&
2118.77$\pm$	65.65&
1771.62$\pm$ 41.05\\
\hline
 
\end{tabular}
\caption{Performance by RMSE (mean) and 95 \% confidence interval($\pm$) of 30 experimental runs for Anand Vihar.}
\label{tab:rmse_anand}
\end{table*}



\begin{table*}[htbp]
\centering
 \small 
\begin{tabular}{lllll}
\hline
 &  BD-LSTM  & BD-LSTM* & UBD-LSTM& SBD-LSTM \\
\hline
\hline
		
Train &      
1949.35$\pm$31.17	&
2027.01$\pm$33.72	&
2081.15$\pm$ 74.34&
1708.78$\pm$ 35.58\\
Test &      
1833.16$\pm$43.64	&
1732.85$\pm$40.63	&
1842.50$\pm$ 71.36&
1281.22$\pm$ 35.57\\
Step-1 &      
1100.56$\pm$26.64	&
1017.65$\pm$22.34	&
297.75$\pm$ 25.75&
683.00$\pm$29.18 \\
Step-2 &      
1385.24$\pm$37.23	&
1253.59$\pm$27.32	&
1297.76$\pm$ 36.08&
937.01$\pm$23.71 \\
Step-3 &      
1651.61$\pm$41.44	&
1534.16$\pm$24.59	&
1591.85$\pm$ 40.72&
1095.53$\pm$ 36.70\\
Step-4 &      
1721.90$\pm$37.51	&
1659.84$\pm$29.50	&
1834.38$\pm$ 38.84&
1173.04$\pm$ 33.13\\
Step-5 &      
1875.45$\pm$38.82	&
1835.25$\pm$25.71	&
2163.71$\pm$ 52.14&
1249.93$\pm$ 38.67\\
Step-6 &      
1919.22$\pm$45.39	&
1880.20$\pm$26.49	&
2323.89$\pm$63.87 &
1384.16$\pm$ 37.51\\
Step-7 &      
2027.93$\pm$60.23	&
1941.07$\pm$32.05	&
2479.18$\pm$ 71.26&
1465.01$\pm$ 34.17\\
Step 8 &      
2160.84$\pm$67.78	&
2040.73$\pm$42.58	&
2252.38$\pm$ 61.94&
1538.47$\pm$ 35.68\\
Step 9 &      
2235.11$\pm$72.46	&
2075.96$\pm$44.83	&
2112.08$\pm$49.15 &
1618.90$\pm$ 32.90\\
Step 10 &      
2253.73$\pm$70.02	&
2090.06$\pm$47.47	&
2072.01$\pm$43.98 &
1667.10$\pm$37.50 \\
\hline
 
\end{tabular}
\caption{Performance by RMSE (mean) and 95 \% confidence interval($\pm$)  of 30 experimental runs using different BD-LSTM models for Bawana.}
\label{tab:rmse_bawana}
\end{table*}



\begin{table*}[htbp]
\centering
 \small 
\begin{tabular}{lllll}
\hline
 &  BD-LSTM  & BD-LSTM* & UBD-LSTM& SBD-LSTM \\
\hline
\hline
		
Train &      
1894.97$\pm$35.20	&
1867.04$\pm$34.88	&
1837.32$\pm$63.35 &
1394.28$\pm$ 25.60\\
Test &      
1707.09$\pm$34.12	&
1655.13$\pm$34.73	&
1871.76$\pm$ 68.03&
1346.61$\pm$41.85 \\
Step-1 &      
1082.65$\pm$35.59	&
990.05$\pm$24.93	&
376.30$\pm$ 46.39&
661.50$\pm$26.35 \\
Step-2 &      
1412.94$\pm$37.73	&
1312.39$\pm$28.11	&
1368.83$\pm$ 51.53&
1009.89$\pm$22.47 \\
Step-3 &      
1595.74$\pm$	42.79&
1541.37$\pm$36.48	&
1890.12$\pm$ 98.02&
1164.77$\pm$ 26.02\\
Step-4 &      
1702.60$\pm$51.73	&
1616.10$\pm$36.87	&
2030.47$\pm$ 92.59&
1165.97$\pm$ 26.92\\
Step-5 &      
1804.52$\pm$52.03	&
1728.34$\pm$42.14	&
2224.31$\pm$ 92.41&
1414.22$\pm$ 42.56\\
Step-6 &      
1841.11$\pm$57.09	&
1788.62$\pm$41.66	&
2272.49$\pm$ 102.77&
1462.59$\pm$ 45.03\\
Step-7 &      
1894.75$\pm$75.12	&
1816.18$\pm$38.03	&
2236.76$\pm$ 90.09&
1411.73$\pm$  54.63\\
Step 8 &      
1946.97$\pm$	70.98&
1929.45$\pm$41.18	&
2143.55$\pm$67.55 &
1743.37$\pm$ 83.88\\
Step 9 &      
1910.16$\pm$56.44	&
1953.05$\pm$36.84	&
2165.78$\pm$ 79.03&
1786.94$\pm$94.11 \\
Step 10 &      
1879.49$\pm$47.86	&
1875.75$\pm$37.62	&
2008.97$\pm$ 57.63&
1645.09$\pm$86.99 \\
\hline
 
\end{tabular}
\caption{RMSE (mean) and 95 \% confidence interval($\pm$)  of 30 experimental runs using different BD-LSTM models for DTU.}
\label{tab:rmse_dtu}
\end{table*}



\begin{table*}[htbp]
\centering
 \small 
\begin{tabular}{lllll}
\hline
 &  BD-LSTM  & BD-LSTM* & UBD-LSTM& SBD-LSTM \\
\hline
\hline
		
Train &      
1820.40$\pm$41.58	&
1779.95$\pm$33.98	&
1994.93$\pm$ 69.00&
1434.41$\pm$31.50 \\
Test &      
1643.59$\pm$40.90	&
1696.24$\pm$42.56	&
1742.39$\pm$59.96 &
1706.87$\pm$ 55.33\\
Step-1 &      
782.12$\pm$	14.33&
792.82$\pm$	12.90&
302.54$\pm$23.65 &
775.95$\pm$21.48 \\
Step-2 &      
1352.21$\pm$28.23	&
1392.97$\pm$24.17	&
1461.58$\pm$ 34.56&
1139.47$\pm$ 32.01\\
Step-3 &      
1548.57$\pm$32.00	&
1612.39$\pm$34.50	&
1640.58$\pm$ 37.38&
1398.11$\pm$ 54.94\\
Step-4 &      
1618.60$\pm$26.98	&
1643.44$\pm$31.46	&
1847.20$\pm$ 53.83&
1493.66$\pm$ 47.42\\
Step-5 &      
1689.86$\pm$31.89	&
1764.74$\pm$31.53	&
2012.07$\pm$57.44 &
1785.54$\pm$ 53.64\\
Step-6 &      
1752.82$\pm$27.40	&
1844.85$\pm$43.18	&
2038.00$\pm$ 47.06&
1936.81$\pm$ 61.08\\
Step-7 &      
1775.63$\pm$40.77	&
1881.67$\pm$50.28	&
2054.06$\pm$41.14 &
2013.61$\pm$ 62.70\\
Step 8 &      
1920.66$\pm$53.87	&
2002.18$\pm$56.76	&
2058.89$\pm$ 41.50&
2180.80$\pm$66.23 \\
Step 9 &      
2010.76$\pm$52.36	&
2061.93$\pm$56.88	&
2019.43$\pm$ 34.69&
2251.07$\pm$ 68.98\\
Step 10 &      
1984.70$\pm$47.81	&
1965.43$\pm$50.60	&
1989.56$\pm$ 40.73&
2093.66$\pm$ 63.69\\
\hline
 
\end{tabular}
\caption{RMSE (mean) and 95 \% confidence interval($\pm$)  of 30 experimental runs using different BD-LSTM models for Vivek Vihar.}
\label{tab:rmse_vivek}
\end{table*}


 \section{Discussion}
 

 In the respective experiments, we   evaluated the performance of different models on the time-series problem and  studied the effect of COVID-19 lockdown on prediction performance of different models for different seasons.
We find that BD-LSTM is our best performer in terms of both train and test performance with better generalization when compared to other LSTM based models  for Anand Vihar. This could be due to the BD-LSTM architecture which features both forward and backward information for a sequence at each time-step using two LSTM layers \cite{graves2005}. We find that BD-LSTM performs the best for larger time steps (6-10) with better generalization as compared to other models   which indicates that BD-LSTM  captures  better information for long-term ahead prediction. This could be due to the combination of memory cells with two LSTM layers which are able to capture salient features in a temporal sequence when compared to canonical LSTM models. Although BD-LSTM based models have been mainly used for different language modelling, we find them promising for long-term time-series prediction problems.

 In the respective experiments, we also found that BD-LSTM and BD-LSTM* perform well in predicting future PM2.5 values for different monitoring stations which indicates that they  are not affected much due to COVID-19 lockdown phase. UBD-LSTM has the highest RMSE  and confidence interval for test datasets  which indicates that it is not robust. The poor performance  which could be due to the fact that UBD-LSTM uses only one features to predict future steps in contrast to multivariate models which uses a set of features as input that  helps with additional information for predicting future steps. The seasonal model (SBD-LSTM)  performs  slightly better   on test dataset for Anand Vihar, Bawana and DTU along with good generalization performance which indicates that the model is robust to the effects of COVID-19 lockdown period.  It is important to evaluate the effect on performance of different models due to seasonal decrease of PM2.5 values during COVID-19 lockdown period when compared to previous years. We find that different LSTM models with different training strategies are quite robust in modeling the effects of COVOD-19 lockdown for both during and partial lockdown periods.     

 Recent focus has been on studying  effects of the lock downs imposed in different cities of India on the spatial patterns of  air quality both during pre-lockdown and during-lockdown phases. The results have demonstrated a direct implication towards the decrease in the concentration of selected pollutants such as PM2.5 and PM10, with maximum decrease(more than 50\%) in comparison to the pre-lockdown phase \cite{mahato2020effect}. There has also been a significant decrease in the concentration of Nitrogen Dioxide(NO2) which has been a major pollutant in reputed cities of India such as Delhi and Mumbai \cite{shehzad2020impact}. There is a positive correlation between PM2.5 concentration and COVID-19   for countries such as India  and Pakistan  \cite{sipra2020can}. Apart from the major air pollutants, there has been a significant decrease in aerosol optical depth (AOD) and lightning activities in many urban and mining regions of India \cite{ranjan2020effect,chowdhuri2020significant}.

 There is a large effort by the Indian government to move towards clean energy with a massive investment in solar and wind energy which can have benefits \cite{goel2016solar,chaurasiya2019wind}. Moreover, India also has plans to go fully electric in the transportation sector by 2030 \cite{vidhi2018review}. Currently, India has began manufacturing  electric vehicles  with plans to have fully electric rail network by 2024 \cite{gupta2020sustainable}. Renewable energy is expected to rise from 27 \% of total energy demand in 2014 to around 43 \% by 2040 \cite{sarangi2018green}. 
In many developing countries, it is clear that an increase of population density increases the air pollution; however, this can change with type of energy used. In a recent study, in the case of China, it has been shown that  the   increase in population density will reduces air pollution  which has been due to clean energy and public transportation \cite{CHEN2020}. A study on the effects of meteorological conditions and air pollution on COVID-19 transmission from 219 Chinese cities found that air pollution indicators were positively correlated with new confirmed cases and increase in cases were linked with air-quality index  \cite{ZHANG2020}. Such study in case of India would be needed and prediction of air-quality indicators could help in preventative measures.

In future work, Bayesian deep learning methods can be used to provide robust uncertainty quantification in predictions which can extend Bayesian neural networks used for time series prediction \cite{chandra2019langevin}. Other learning strategies such as multi-task and transfer learning in conjunction with Bayesian inference can be used to develop improved models   \cite{chandra2020bayesian,chandra2017comtltime} which can take into account existing deep learning models for COVID-19 infections in India \cite{chandra2021deepcovid}. 
Moreover, we envision the application of the proposed framework in other parts of India and rest of the world which has decline in air quality. A web-based framework can be implemented by the respective authorities that can be used to provide proper weekly and monthly planning, which could involve the way traffic is managed for  different seasons. The methodology can also be extended for air quality prediction in relation to forest fires around the world.



\section{Conclusion and Future Work} 
 
 In this paper, we applied deep learning via different LSTM models for univariate and multivariate modelling for short-term and long term air quality prediction taking into account the four base-stations from Delhi, India. Although we selected a subset of the base stations, the methodology can be applied to rest of the base stations in Delhi or in other parts of the world.  Our results show that multivariate bi-directional LSTM model shows best performance, and rest of the  LSTM models have certain strengths and limitations which need to be evaluated prior to developing a system that provides rigours uncertainty quantification in predictions. We also found that COVID-19 had a significant  effect on the air quality  during full lockdown implemented  for few months and afterwards, there was unprecedented growth of poor air quality which has a seasonal effect as compared to previous years. We provide open source software framework and open data which can be used for further verification and also application for studying or developing prediction models in other places.

\section{Software and Data}
 We provide Python based open source implementation along with the data for different experiments and design of respective method for further research \footnote{\url{https://github.com/sydney-machine-learning/airpollution_deeplearning}}.
\bibliographystyle{IEEEtran}
\bibliography{usyd,Chandra-Rohitash,Bays,2018,aicrg,rr,sample,sample_,2019June,covid,august2020}

\begin{thebibliography}{100}
\providecommand{\url}[1]{#1}
\csname url@samestyle\endcsname
\providecommand{\newblock}{\relax}
\providecommand{\bibinfo}[2]{#2}
\providecommand{\BIBentrySTDinterwordspacing}{\spaceskip=0pt\relax}
\providecommand{\BIBentryALTinterwordstretchfactor}{4}
\providecommand{\BIBentryALTinterwordspacing}{\spaceskip=\fontdimen2\font plus
\BIBentryALTinterwordstretchfactor\fontdimen3\font minus
  \fontdimen4\font\relax}
\providecommand{\BIBforeignlanguage}[2]{{%
\expandafter\ifx\csname l@#1\endcsname\relax
\typeout{** WARNING: IEEEtran.bst: No hyphenation pattern has been}%
\typeout{** loaded for the language `#1'. Using the pattern for}%
\typeout{** the default language instead.}%
\else
\language=\csname l@#1\endcsname
\fi
#2}}
\providecommand{\BIBdecl}{\relax}
\BIBdecl

\bibitem{MaxRoser2013}
H.~R. Max~Roser and E.~Ortiz-Ospina, ``World population growth,'' \emph{Our
  World in Data}, 2013, https://ourworldindata.org/world-population-growth.

\bibitem{desa2015united}
U.~Desa, ``United nations department of economic and social affairs, population
  division. world population prospects: The 2015 revision, key findings and
  advance tables,'' in \emph{Technical Report: Working Paper No. ESA/P/WP.
  241}, 2015.

\bibitem{taylor2004world}
P.~J. Taylor and B.~Derudder, \emph{World city network: a global urban
  analysis}.\hskip 1em plus 0.5em minus 0.4em\relax Routledge, 2004.

\bibitem{who2016}
WHO, ``Air pollution levels rising in many of the world’s poorest cities,''
  2016.

\bibitem{unep2016}
UNEP, ``Status of fuel quality and vehicle emission standards latin america,''
  2016.

\bibitem{harlan2011climate}
S.~L. Harlan and D.~M. Ruddell, ``Climate change and health in cities: impacts
  of heat and air pollution and potential co-benefits from mitigation and
  adaptation,'' \emph{Current Opinion in Environmental Sustainability}, vol.~3,
  no.~3, pp. 126--134, 2011.

\bibitem{gupta2006satellite}
P.~Gupta, S.~A. Christopher, J.~Wang, R.~Gehrig, Y.~Lee, and N.~Kumar,
  ``Satellite remote sensing of particulate matter and air quality assessment
  over global cities,'' \emph{Atmospheric Environment}, vol.~40, no.~30, pp.
  5880--5892, 2006.

\bibitem{zhang2015fine}
Y.-L. Zhang and F.~Cao, ``Fine particulate matter (pm 2.5) in china at a city
  level,'' \emph{Scientific reports}, vol.~5, p. 14884, 2015.

\bibitem{liu2015transparent}
C.~Liu, P.-C. Hsu, H.-W. Lee, M.~Ye, G.~Zheng, N.~Liu, W.~Li, and Y.~Cui,
  ``Transparent air filter for high-efficiency pm 2.5 capture,'' \emph{Nature
  communications}, vol.~6, no.~1, pp. 1--9, 2015.

\bibitem{franklin2007association}
M.~Franklin, A.~Zeka, and J.~Schwartz, ``Association between pm 2.5 and
  all-cause and specific-cause mortality in 27 us communities,'' \emph{Journal
  of exposure science \& environmental epidemiology}, vol.~17, no.~3, pp.
  279--287, 2007.

\bibitem{lelieveld2015contribution}
J.~Lelieveld, J.~S. Evans, M.~Fnais, D.~Giannadaki, and A.~Pozzer, ``The
  contribution of outdoor air pollution sources to premature mortality on a
  global scale,'' \emph{Nature}, vol. 525, no. 7569, pp. 367--371, 2015.

\bibitem{indiapop}
\BIBentryALTinterwordspacing
``Unique identification authority of {India},'' May 2020, [Online; accessed
  16-July-2020]. [Online]. Available:
  \url{https://uidai.gov.in/images/state-wise-aadhaar-saturation.pdf}
\BIBentrySTDinterwordspacing

\bibitem{weforum2020}
\BIBentryALTinterwordspacing
``6 of the world’s 10 most polluted cities are in {India},'' August 2020,
  [Online; accessed 22-August-2020]. [Online]. Available:
  \url{https://www.weforum.org/agenda/2020/03/6-of-the-world-s-10-most-polluted-cities-are-in-india/}
\BIBentrySTDinterwordspacing

\bibitem{who2020}
\BIBentryALTinterwordspacing
``7 million premature deaths annually linked to air pollution,'' August 2020,
  [Online; accessed 22-August-2020]. [Online]. Available:
  \url{https://www.who.int/mediacentre/news/releases/2014/air-pollution/en/}
\BIBentrySTDinterwordspacing

\bibitem{apte2018ambient}
J.~S. Apte, M.~Brauer, A.~J. Cohen, M.~Ezzati, and C.~A. Pope~III, ``Ambient
  pm2. 5 reduces global and regional life expectancy,'' \emph{Environmental
  Science \& Technology Letters}, vol.~5, no.~9, pp. 546--551, 2018.

\bibitem{cusworth2018quantifying}
D.~H. Cusworth, L.~J. Mickley, M.~P. Sulprizio, T.~Liu, M.~E. Marlier, R.~S.
  DeFries, S.~K. Guttikunda, and P.~Gupta, ``Quantifying the influence of
  agricultural fires in northwest india on urban air pollution in delhi,
  india,'' \emph{Environmental Research Letters}, vol.~13, no.~4, p. 044018,
  2018.

\bibitem{hogland2000assessment}
W.~Hogland and J.~Stenis, ``Assessment and system analysis of industrial waste
  management,'' \emph{Waste Management}, vol.~20, no.~7, pp. 537--543, 2000.

\bibitem{le2020spatiotemporal}
V.-D. Le, T.-C. Bui, and S.-K. Cha, ``Spatiotemporal deep learning model for
  citywide air pollution interpolation and prediction,'' in \emph{2020 IEEE
  International Conference on Big Data and Smart Computing (BigComp)}.\hskip
  1em plus 0.5em minus 0.4em\relax IEEE, 2020, pp. 55--62.

\bibitem{li2016deep}
X.~Li, L.~Peng, Y.~Hu, J.~Shao, and T.~Chi, ``Deep learning architecture for
  air quality predictions,'' \emph{Environmental Science and Pollution
  Research}, vol.~23, no.~22, pp. 22\,408--22\,417, 2016.

\bibitem{bui2018deep}
T.-C. Bui, V.-D. Le, and S.-K. Cha, ``A deep learning approach for forecasting
  air pollution in south korea using lstm,'' \emph{arXiv preprint
  arXiv:1804.07891}, 2018.

\bibitem{grell2005}
G.~Grell, S.~Peckham, R.~Schmitz, S.~McKeen, G.~Frost, W.~Skamarock, and
  B.~Eder, ``Fully coupled “online” chemistry in the wrf model,''
  \emph{Atmospheric Environment}, vol.~39, pp. 6957--6975, 12 2005.

\bibitem{emmons2009}
L.~Emmons, S.~Walters, P.~Hess, J.-F. Lamarque, G.~Pfister, F.~D, C.~Granier,
  A.~Guenther, K.~D, T.~Laepple, O.~J, X.~Tie, T.~G, C.~Wiedinmyer,
  S.~Baughcum, and S.~Kloster, ``Description and evaluation of the model for
  ozone and related chemical tracers, version 4 (mozart-4),''
  \emph{Geoscientific Model Development Discussions}, vol.~3, 01 2009.

\bibitem{karimian2016hamed}
H.~Karimian, Q.~Li, C.~LI, L.~Jin, J.~Fan, and Y.~Li, ``An improved method for
  monitoring fine particulate matter mass concentrations via satellite remote
  sensing,'' \emph{Aerosol and Air Quality Research}, vol.~16, pp. 1081--1092,
  01 2016.

\bibitem{Yuan2017PredictingTA}
Z.~Yuan, X.~Zhou, T.~Yang, and J.~Tamerius, ``Predicting traffic accidents
  through heterogeneous urban data : A case study,'' 2017.

\bibitem{10.1145/1390156.1390177}
\BIBentryALTinterwordspacing
R.~Collobert and J.~Weston, ``A unified architecture for natural language
  processing: Deep neural networks with multitask learning,'' in
  \emph{Proceedings of the 25th International Conference on Machine Learning},
  ser. ICML ’08.\hskip 1em plus 0.5em minus 0.4em\relax New York, NY, USA:
  Association for Computing Machinery, 2008, p. 160–167. [Online]. Available:
  \url{https://doi.org/10.1145/1390156.1390177}
\BIBentrySTDinterwordspacing

\bibitem{fan2008}
J.~Fan, Y.~Gao, and H.~Luo, ``Integrating concept ontology and multitask
  learning to achieve more effective classifier training for multilevel image
  annotation,'' \emph{IEEE transactions on image processing : a publication of
  the IEEE Signal Processing Society}, vol.~17, pp. 407--26, 04 2008.

\bibitem{inproceedings}
C.~Widmer, J.~Leiva, Y.~Altun, and G.~Rätsch, ``Leveraging sequence
  classification by taxonomy-based multitask learning,'' vol. 6044, 04 2010,
  pp. 522--534.

\bibitem{lindbeck2000}
A.~Lindbeck and D.~J. Snower, ``Multitask learning and the reorganization of
  work: from tayloristic to holistic organization,'' \emph{Journal of labor
  economics}, vol.~18, no.~3, pp. 353--376, 2000.

\bibitem{patel2017analytic}
C.~J. Patel, ``Analytic complexity and challenges in identifying mixtures of
  exposures associated with phenotypes in the exposome era,'' \emph{Current
  epidemiology reports}, vol.~4, no.~1, pp. 22--30, 2017.

\bibitem{schmidhuber2015deep}
J.~Schmidhuber, ``Deep learning in neural networks: An overview,'' \emph{Neural
  networks}, vol.~61, pp. 85--117, 2015.

\bibitem{elman_Zipser1988}
\BIBentryALTinterwordspacing
J.~L. Elman and D.~Zipser, ``Learning the hidden structure of speech,''
  \emph{The Journal of the Acoustical Society of America}, vol.~83, no.~4, pp.
  1615--1626, 1988. [Online]. Available:
  \url{http://link.aip.org/link/?JAS/83/1615/1}
\BIBentrySTDinterwordspacing

\bibitem{Werbos_1990}
P.~J. Werbos, ``Backpropagation through time: what it does and how to do it,''
  \emph{Proceedings of the IEEE}, vol.~78, no.~10, pp. 1550--1560, 1990.

\bibitem{hochreiter1997long}
S.~Hochreiter and J.~Schmidhuber, ``Long short-term memory,'' \emph{Neural
  computation}, vol.~9, no.~8, pp. 1735--1780, 1997.

\bibitem{ChandraTNNLS2015}
R.~Chandra, ``Competition and collaboration in cooperative coevolution of
  {Elman} recurrent neural networks for time-series prediction,'' \emph{Neural
  Networks and Learning Systems, IEEE Transactions on}, vol.~26, pp.
  3123--3136, 2015.

\bibitem{hochreiter1998vanishing}
S.~Hochreiter, ``The vanishing gradient problem during learning recurrent
  neural nets and problem solutions,'' \emph{International Journal of
  Uncertainty, Fuzziness and Knowledge-Based Systems}, vol.~6, no.~02, pp.
  107--116, 1998.

\bibitem{bengio1994learning}
Y.~Bengio, P.~Simard, and P.~Frasconi, ``Learning long-term dependencies with
  gradient descent is difficult,'' \emph{Neural Networks, IEEE Transactions
  on}, vol.~5, no.~2, pp. 157--166, 1994.

\bibitem{chung2014empirical}
J.~Chung, C.~Gulcehre, K.~Cho, and Y.~Bengio, ``Empirical evaluation of gated
  recurrent neural networks on sequence modeling,'' \emph{arXiv preprint
  arXiv:1412.3555}, 2014.

\bibitem{cho2014learning}
K.~Cho, B.~Van~Merri{\"e}nboer, C.~Gulcehre, D.~Bahdanau, F.~Bougares,
  H.~Schwenk, and Y.~Bengio, ``Learning phrase representations using rnn
  encoder-decoder for statistical machine translation,'' \emph{arXiv preprint
  arXiv:1406.1078}, 2014.

\bibitem{650093}
M.~{Schuster} and K.~K. {Paliwal}, ``Bidirectional recurrent neural networks,''
  \emph{IEEE Transactions on Signal Processing}, vol.~45, no.~11, pp.
  2673--2681, 1997.

\bibitem{10.1016/j.neunet.2005.06.042}
\BIBentryALTinterwordspacing
A.~Graves and J.~Schmidhuber, ``2005 special issue: Framewise phoneme
  classification with bidirectional lstm and other neural network
  architectures,'' \emph{Neural Netw.}, vol.~18, no. 5–6, p. 602–610, Jun.
  2005. [Online]. Available: \url{https://doi.org/10.1016/j.neunet.2005.06.042}
\BIBentrySTDinterwordspacing

\bibitem{Gorbalenya2020species}
A.~E. Gorbalenya, S.~C. Baker, R.~S. Baric, R.~J. de~Groot, C.~Drosten, A.~A.
  Gulyaeva, B.~L. Haagmans, C.~Lauber, A.~M. Leontovich, B.~W. Neuman,
  D.~Penzar, L.~L. M.~P. Stanley~Perlman10, D.~V. Samborskiy, I.~A. Sidorov,
  I.~Sola, and J.~Ziebuhr, ``The species severe acute respiratory
  syndrome-related coronavirus: classifying 2019-ncov and naming it
  sars-cov-2,'' \emph{Nature Microbiology}, vol.~5, no.~4, p. 536, 2020.

\bibitem{monteil2020inhibition}
V.~Monteil, H.~Kwon, P.~Prado, A.~Hagelkr{\"u}ys, R.~A. Wimmer, M.~Stahl,
  A.~Leopoldi, E.~Garreta, C.~H. Del~Pozo, F.~Prosper \emph{et~al.},
  ``Inhibition of sars-cov-2 infections in engineered human tissues using
  clinical-grade soluble human ace2,'' \emph{Cell}, 2020.

\bibitem{world2020coronavirus}
W.~H. Organization \emph{et~al.}, ``Coronavirus disease 2019 ( {COVID-19}):
  situation report, 72,'' 2020.

\bibitem{cucinotta2020declares}
D.~Cucinotta and M.~Vanelli, ``{WHO} declares {COVID-19} a pandemic,''
  \emph{Acta bio-medica: Atenei Parmensis}, vol.~91, no.~1, pp. 157--160, 2020.

\bibitem{atkeson2020will}
A.~Atkeson, ``What will be the economic impact of {COVID-19} in the {US}? rough
  estimates of disease scenarios,'' National Bureau of Economic Research, Tech.
  Rep., 2020.

\bibitem{fernandes2020economic}
N.~Fernandes, ``Economic effects of coronavirus outbreak (covid-19) on the
  world economy,'' \emph{Available at SSRN 3557504}, 2020.

\bibitem{maliszewska2020potential}
M.~Maliszewska, A.~Mattoo, and D.~Van Der~Mensbrugghe, ``The potential impact
  of {COVID-19} on gdp and trade: A preliminary assessment,'' 2020.

\bibitem{kerimray2020assessing}
A.~Kerimray, N.~Baimatova, O.~P. Ibragimova, B.~Bukenov, B.~Kenessov,
  P.~Plotitsyn, and F.~Karaca, ``Assessing air quality changes in large cities
  during {COVID-19} lockdowns: The impacts of traffic-free urban conditions in
  almaty, kazakhstan,'' \emph{Science of the Total Environment}, p. 139179,
  2020.

\bibitem{dantas2020impact}
G.~Dantas, B.~Siciliano, B.~B. Fran{\c{c}}a, C.~M. da~Silva, and G.~Arbilla,
  ``The impact of {COVID-19} partial lockdown on the air quality of the city of
  rio de janeiro, brazil,'' \emph{Science of the Total Environment}, vol. 729,
  p. 139085, 2020.

\bibitem{yongjian2020association}
Z.~Yongjian, X.~Jingu, H.~Fengming, and C.~Liqing, ``Association between
  short-term exposure to air pollution and {COVID-19} infection: Evidence from
  china,'' \emph{Science of the total environment}, p. 138704, 2020.

\bibitem{wang2020severe}
P.~Wang, K.~Chen, S.~Zhu, P.~Wang, and H.~Zhang, ``Severe air pollution events
  not avoided by reduced anthropogenic activities during {COVID-19} outbreak,''
  \emph{Resources, Conservation and Recycling}, vol. 158, p. 104814, 2020.

\bibitem{bbcDelhi2020}
\BIBentryALTinterwordspacing
``India coronavirus: Can the covid-19 lockdown spark a clean air movement?''
  August 2020, [Online; accessed 24-August-2020]. [Online]. Available:
  \url{https://www.bbc.com/news/world-asia-india-52313972}
\BIBentrySTDinterwordspacing

\bibitem{BAO202013}
R.~Bao and A.~Zhang, ``Does lockdown reduce air pollution? evidence from 44
  cities in northern {China},'' \emph{Science of The Total Environment}, vol.
  731, p. 139052, 2020.

\bibitem{li2020air}
L.~Li, Q.~Li, L.~Huang, Q.~Wang, A.~Zhu, J.~Xu, Z.~Liu, H.~Li, L.~Shi, R.~Li
  \emph{et~al.}, ``Air quality changes during the {COVID-19} lockdown over the
  yangtze river delta region: An insight into the impact of human activity
  pattern changes on air pollution variation,'' \emph{Science of The Total
  Environment}, p. 139282, 2020.

\bibitem{dutheil2020covid}
F.~Dutheil, J.~S. Baker, and V.~Navel, ``{COVID-19} as a factor influencing air
  pollution?'' \emph{Environmental Pollution (Barking, Essex: 1987)}, vol. 263,
  p. 114466, 2020.

\bibitem{tosepu2020correlation}
R.~Tosepu, J.~Gunawan, D.~S. Effendy, H.~Lestari, H.~Bahar, P.~Asfian
  \emph{et~al.}, ``Correlation between weather and {COVID-19} pandemic in
  jakarta, indonesia,'' \emph{Science of The Total Environment}, p. 138436,
  2020.

\bibitem{ma2020effects}
Y.~Ma, Y.~Zhao, J.~Liu, X.~He, B.~Wang, S.~Fu, J.~Yan, J.~Niu, J.~Zhou, and
  B.~Luo, ``Effects of temperature variation and humidity on the death of
  {COVID-19 in Wuhan, China},'' \emph{Science of The Total Environment}, p.
  138226, 2020.

\bibitem{wang2003}
W.~Wang, Z.~Xu, and W.-Z. Lu, ``Three improved neural network models for air
  quality forecasting,'' \emph{Engineering Computations}, vol.~20, pp.
  192--210, 03 2003.

\bibitem{corani2005}
G.~Corani, ``Air quality prediction in milan: Feed-forward neural networks,
  pruned neural networks and lazy learning,'' \emph{Ecological Modelling}, pp.
  513--529, 07 2005.

\bibitem{Peng2015AirQP}
H.~Peng, ``Air quality prediction by machine learning methods,'' 2015.

\bibitem{fu2015}
M.~Fu, W.~Wang, Z.~Le, and M.~Safaei~khorram, ``Prediction of particular matter
  concentrations by developed feed-forward neural network with rolling
  mechanism and gray model,'' \emph{Neural Computing and Applications}, 02
  2015.

\bibitem{CHANG20201451}
\BIBentryALTinterwordspacing
Y.-S. Chang, H.-T. Chiao, S.~Abimannan, Y.-P. Huang, Y.-T. Tsai, and K.-M. Lin,
  ``An lstm-based aggregated model for air pollution forecasting,''
  \emph{Atmospheric Pollution Research}, vol.~11, no.~8, pp. 1451 -- 1463,
  2020. [Online]. Available:
  \url{http://www.sciencedirect.com/science/article/pii/S1309104220301215}
\BIBentrySTDinterwordspacing

\bibitem{xiao2018}
D.~Xiao, J.~Zheng, C.~Pain, and I.~Navon, ``Machine learning-based rapid
  response tools for regional air pollution modelling,'' \emph{Atmospheric
  Environment}, vol. 199, 11 2018.

\bibitem{zhu2018}
D.~Zhu, C.~Cai, T.~Yang, and X.~Zhou, ``A machine learning approach for air
  quality prediction: Model regularization and optimization,'' \emph{Big Data
  and Cognitive Computing}, vol.~2, p.~5, 02 2018.

\bibitem{kim2014prenatal}
E.~Kim, H.~Park, Y.-C. Hong, M.~Ha, Y.~Kim, B.-N. Kim, Y.~Kim, Y.-M. Roh, B.-E.
  Lee, J.-M. Ryu \emph{et~al.}, ``Prenatal exposure to pm10 and no2 and
  children's neurodevelopment from birth to 24 months of age: Mothers and
  children's environmental health (moceh) study,'' \emph{Science of the Total
  Environment}, vol. 481, pp. 439--445, 2014.

\bibitem{HUANG2020}
Y.~Huang, Y.~Wu, and W.~Zhang, ``Comprehensive identification and isolation
  policies have effectively suppressed the spread of {COVID-19},'' \emph{Chaos,
  Solitons\& Fractals}, vol. 139, p. 110041, 2020.

\bibitem{MA2020}
X.~Ma, I.~Longley, J.~Gao, and J.~Salmond, ``Assessing schoolchildren's
  exposure to air pollution during the daily commute - a systematic review,''
  \emph{Science of The Total Environment}, vol. 737, p. 140389, 2020.

\bibitem{LIU2020}
K.~Liu, B.-Y. Yang, Y.~Guo, M.~S. Bloom, S.~C. Dharmage, L.~D. Knibbs,
  J.~Heinrich, A.~Leskinen, S.~Lin, L.~Morawska, B.~Jalaludin, I.~Markevych,
  P.~Jalava, M.~Komppula, Y.~Yu, M.~Gao, Y.~Zhou, H.-Y. Yu, L.-W. Hu, X.-W.
  Zeng, and G.-H. Dong, ``The role of influenza vaccination in mitigating the
  adverse impact of ambient air pollution on lung function in children: New
  insights from the seven northeastern cities study in china,''
  \emph{Environmental Research}, vol. 187, p. 109624, 2020.

\bibitem{WANG2020}
Z.~Wang, W.~Wei, and F.~Zheng, ``Effects of industrial air pollution on the
  technical efficiency of agricultural production: Evidence from china,''
  \emph{Environmental Impact Assessment Review}, vol.~83, p. 106407, 2020.

\bibitem{XU2020x}
M.~Xu, Y.~Wang, and Y.~Tu, ``Uncovering the invisible effect of air pollution
  on stock returns: A moderation and mediation analysis,'' \emph{Finance
  Research Letters}, p. 101646, 2020.

\bibitem{UNworldcities2016}
\BIBentryALTinterwordspacing
``The worlds cities in 2016: data booklet,'' June 2016, [United Nations,
  Online; accessed 24-August-2020]. [Online]. Available:
  \url{https://www.un.org/en/development/desa/population/publications/index.asp}
\BIBentrySTDinterwordspacing

\bibitem{indiatimesvehicles2020}
\BIBentryALTinterwordspacing
``Usual suspects: Vehicles, industrial emissions behind foul play,'' August
  2018, [Online; accessed 24-August-2020]. [Online]. Available:
  \url{https://timesofindia.indiatimes.com/city/delhi/usual-suspects-vehicles-industrial-emissions-behind-foul-play-all-year/articleshow/66228517.cms}
\BIBentrySTDinterwordspacing

\bibitem{delhiJanApril}
\BIBentryALTinterwordspacing
``Delhi breathed easier from january to april,'' June 2017, [Online; accessed
  24-August-2020]. [Online]. Available:
  \url{https://timesofindia.indiatimes.com/city/delhi/delhi-breathed-easier-from-january-to-april/articleshow/59011204.cms}
\BIBentrySTDinterwordspacing

\bibitem{delhifeb}
\BIBentryALTinterwordspacing
``Air pollution: {Delhi} enjoys cleanest february in three years,'' February
  2018, [Online; accessed 24-August-2020]. [Online]. Available:
  \url{https://www.hindustantimes.com/delhi-news/air-pollution-delhi-enjoys-cleanest-february-in-three-years/story-SANlmslHev8ifFgZbh3WXI.html}
\BIBentrySTDinterwordspacing

\bibitem{sulian2013}
R.~Suliankatchi, B.~Nongkynrih, and S.~Gupta, ``Air pollution in delhi: Its
  magnitude and effects on health,'' \emph{Indian Journal of Community
  Medicine}, vol.~38, pp. 4--8, 01 2013.

\bibitem{ab2ef8b1142c4a86ae007857626db575}
A.~Comrie, ``\BIBforeignlanguage{English (US)}{A synoptic climatology of rural
  ozone pollution at three forest sites in pennsylvania},''
  \emph{\BIBforeignlanguage{English (US)}{Atmospheric Environment}}, vol.~28,
  no.~9, pp. 1601--1614, May 1994.

\bibitem{article}
A.~Degaetano and O.~Doherty, ``Temporal, spatial and meteorological variations
  in hourly pm 2.5 concentration extremes in new york city,'' \emph{Atmospheric
  Environment - ATMOS ENVIRON}, vol.~38, pp. 1547--1558, 04 2004.

\bibitem{10.1175/1520-0450(1994)033<1182:AACSDT>2.0.CO;2}
\BIBentryALTinterwordspacing
B.~K. Eder, J.~M. Davis, and P.~Bloomfield, ``{An Automated Classification
  Scheme Designed to Better Elucidate the Dependence of Ozone on
  Meteorology},'' \emph{Journal of Applied Meteorology}, vol.~33, no.~10, pp.
  1182--1199, 10 1994. [Online]. Available:
  \url{https://doi.org/10.1175/1520-0450(1994)033<1182:AACSDT>2.0.CO;2}
\BIBentrySTDinterwordspacing

\bibitem{1997AtmEn..31..869Z}
M.~P. {Zelenka}, ``{An analysis of the meteorological parameters affecting
  ambient concentrations of acid aerosols in Uniontown, Pennsylvania},''
  \emph{Atmospheric Environment}, vol.~31, no.~6, pp. 869--878, Jan. 1997.

\bibitem{watson1988atmospheric}
A.~Y. Watson, R.~R. Bates, and D.~Kennedy, ``Atmospheric transport and
  dispersion of air pollutants associated with vehicular emissions,'' in
  \emph{Air Pollution, the Automobile, and Public Health}.\hskip 1em plus 0.5em
  minus 0.4em\relax National Academies Press (US), 1988.

\bibitem{natsadorj2003}
L.~Natsagdorj, D.~Jugder, and Y.~Chung, ``Analysis of dust storms observed in
  mongolia during 1937–1999,'' \emph{Atmospheric Environment}, vol.~37, pp.
  1401--1411, 03 2003.

\bibitem{zalakeviciute2018contrasted}
R.~Zalakeviciute, J.~L{\'o}pez-Villada, and Y.~Rybarczyk, ``Contrasted effects
  of relative humidity and precipitation on urban pm2. 5 pollution in high
  elevation urban areas,'' \emph{Sustainability}, vol.~10, no.~6, p. 2064,
  2018.

\bibitem{seinfeld2016}
S.~Seinfeld, J.H.;~Pandis, ``Atmospheric chemistry and physics: From air
  pollution to climate change,'' 2016.

\bibitem{1985AtmEn..19.1525A}
B.~R. {Appel}, Y.~{Tokiwa}, J.~{Hsu}, E.~L. {Kothny}, and E.~{Hahn},
  ``{Visibility as related to atmospheric aerosol constituents},''
  \emph{Atmospheric Environment}, vol.~19, no.~9, pp. 1525--1534, Jan. 1985.

\bibitem{10.1175/1520-0469(1977)034<1149:TIOPOT>2.0.CO;2}
\BIBentryALTinterwordspacing
S.~Twomey, ``{The Influence of Pollution on the Shortwave Albedo of Clouds},''
  \emph{Journal of the Atmospheric Sciences}, vol.~34, no.~7, pp. 1149--1152,
  07 1977. [Online]. Available:
  \url{https://doi.org/10.1175/1520-0469(1977)034<1149:TIOPOT>2.0.CO;2}
\BIBentrySTDinterwordspacing

\bibitem{akbari2002}
H.~Akbari, ``Shade trees reduce building energy use and co2 emissions from
  power plants,'' \emph{Environmental pollution (Barking, Essex : 1987)}, vol.
  116 Suppl 1, pp. S119--26, 02 2002.

\bibitem{CCR:2020}
\BIBentryALTinterwordspacing
G.~o.~I. Central Pollution Control~Board, \emph{CCR}, accessed 20 December,
  2020. [Online]. Available:
  \url{https://app.cpcbccr.com/ccr/#/caaqm-dashboard/caaqm-landing}
\BIBentrySTDinterwordspacing

\bibitem{Elman_1990}
J.~L. Elman, ``Finding structure in time,'' \emph{Cognitive Science}, vol.~14,
  pp. 179--211, 1990.

\bibitem{Omlin_thonberetal1996}
C.~W. Omlin, K.~K. Thornber, and C.~L. Giles, ``Fuzzy finite state automata can
  be deterministically encoded into recurrent neural networks,'' \emph{IEEE
  Trans. Fuzzy Syst.}, vol.~6, pp. 76--89, 1998.

\bibitem{Omlin_Giles1992}
C.~W. Omlin and C.~L. Giles, ``Training second-order recurrent neural networks
  using hints,'' in \emph{Proceedings of the Ninth International Conference on
  Machine Learning}.\hskip 1em plus 0.5em minus 0.4em\relax Morgan Kaufmann,
  1992, pp. 363--368.

\bibitem{Giles_etal1999}
C.~L. Giles, C.~Omlin, and K.~K. Thornber, ``Equivalence in knowledge
  representation: Automata, recurrent neural networks, and dynamical fuzzy
  systems,'' \emph{Proceedings of the IEEE}, vol.~87, no.~9, pp. 1623--1640,
  1999.

\bibitem{Hochreiter_1998}
S.~Hochreiter, ``The vanishing gradient problem during learning recurrent
  neural nets and problem solutions,'' \emph{Int. J. Uncertain. Fuzziness
  Knowl.-Based Syst.}, vol.~6, no.~2, pp. 107--116, 1998.

\bibitem{kingma2014adam}
D.~P. Kingma and J.~Ba, ``Adam: {A} method for stochastic optimization,'' in
  \emph{3rd International Conference on Learning Representations, {ICLR} 2015,
  San Diego, CA, USA, May 7-9, 2015, Conference Track Proceedings}, 2015.

\bibitem{schuster1997}
M.~Schuster and K.~Paliwal, ``Bidirectional recurrent neural networks,''
  \emph{Signal Processing, IEEE Transactions on}, vol.~45, pp. 2673 -- 2681, 12
  1997.

\bibitem{graves2005}
A.~Graves and J.~Schmidhuber, ``Framewise phoneme classification with
  bidirectional lstm and other neural network architectures,'' \emph{Neural
  networks : the official journal of the International Neural Network Society},
  vol.~18, pp. 602--10, 07 2005.

\bibitem{Fan2014TTSSW}
Y.~Fan, Y.~Qian, F.-L. Xie, and F.~K. Soong, ``Tts synthesis with bidirectional
  lstm based recurrent neural networks,'' in \emph{INTERSPEECH}, 2014.

\bibitem{graves2013hybrid}
A.~Graves, N.~Jaitly, and A.-r. Mohamed, ``Hybrid speech recognition with deep
  bidirectional lstm,'' in \emph{2013 IEEE workshop on automatic speech
  recognition and understanding}.\hskip 1em plus 0.5em minus 0.4em\relax IEEE,
  2013, pp. 273--278.

\bibitem{ding2018densely}
Z.~Ding, R.~Xia, J.~Yu, X.~Li, and J.~Yang, ``Densely connected bidirectional
  lstm with applications to sentence classification,'' in \emph{CCF
  International Conference on Natural Language Processing and Chinese
  Computing}.\hskip 1em plus 0.5em minus 0.4em\relax Springer, 2018, pp.
  278--287.

\bibitem{NIPS2014_5346}
\BIBentryALTinterwordspacing
I.~Sutskever, O.~Vinyals, and Q.~V. Le, ``Sequence to sequence learning with
  neural networks,'' in \emph{Advances in Neural Information Processing Systems
  27}, Z.~Ghahramani, M.~Welling, C.~Cortes, N.~D. Lawrence, and K.~Q.
  Weinberger, Eds.\hskip 1em plus 0.5em minus 0.4em\relax Curran Associates,
  Inc., 2014, pp. 3104--3112. [Online]. Available:
  \url{http://papers.nips.cc/paper/5346-sequence-to-sequence-learning-with-neural-networks.pdf}
\BIBentrySTDinterwordspacing

\bibitem{wang2016experimental}
T.~Wang, P.~Chen, K.~Amaral, and J.~Qiang, ``An experimental study of lstm
  encoder-decoder model for text simplification,'' \emph{arXiv preprint
  arXiv:1609.03663}, 2016.

\bibitem{zeyer2019comparison}
A.~Zeyer, P.~Bahar, K.~Irie, R.~Schl{\"u}ter, and H.~Ney, ``A comparison of
  transformer and lstm encoder decoder models for asr,'' in \emph{2019 IEEE
  Automatic Speech Recognition and Understanding Workshop (ASRU)}.\hskip 1em
  plus 0.5em minus 0.4em\relax IEEE, 2019, pp. 8--15.

\bibitem{yao2015sequence}
K.~Yao and G.~Zweig, ``Sequence-to-sequence neural net models for
  grapheme-to-phoneme conversion,'' \emph{arXiv preprint arXiv:1506.00196},
  2015.

\bibitem{masum2018}
S.~Masum, Y.~Liu, and J.~Chiverton, \emph{Multi-step Time Series Forecasting of
  Electric Load Using Machine Learning Models}, 01 2018, pp. 148--159.

\bibitem{mahato2020effect}
S.~Mahato, S.~Pal, and K.~G. Ghosh, ``Effect of lockdown amid covid-19 pandemic
  on air quality of the megacity delhi, india,'' \emph{Science of the Total
  Environment}, p. 139086, 2020.

\bibitem{shehzad2020impact}
K.~Shehzad, M.~Sarfraz, and S.~G.~M. Shah, ``The impact of covid-19 as a
  necessary evil on air pollution in india during the lockdown,''
  \emph{Environmental Pollution}, vol. 266, p. 115080, 2020.

\bibitem{sipra2020can}
S.~Sipra, M.~M. Abrar, M.~Iqbal, E.~Haider, and H.~M.~H. Shoukat, ``Can pm2. 5
  pollution worsen the death rate due to covid-19 in india and pakistan?''
  \emph{The Science of the Total Environment}, vol. 742, p. 140557, 2020.

\bibitem{ranjan2020effect}
A.~K. Ranjan, A.~Patra, and A.~Gorai, ``Effect of lockdown due to sars covid-19
  on aerosol optical depth (aod) over urban and mining regions in india,''
  \emph{Science of the Total Environment}, vol. 745, p. 141024, 2020.

\bibitem{chowdhuri2020significant}
I.~Chowdhuri, S.~C. Pal, A.~Saha, R.~Chakrabortty, M.~Ghosh, and P.~Roy,
  ``Significant decrease of lightning activities during covid-19 lockdown
  period over kolkata megacity in india,'' \emph{Science of the Total
  Environment}, vol. 747, p. 141321, 2020.

\bibitem{goel2016solar}
M.~Goel, ``Solar rooftop in india: Policies, challenges and outlook,''
  \emph{Green Energy \& Environment}, vol.~1, no.~2, pp. 129--137, 2016.

\bibitem{chaurasiya2019wind}
P.~K. Chaurasiya, V.~Warudkar, and S.~Ahmed, ``Wind energy development and
  policy in india: A review,'' \emph{Energy Strategy Reviews}, vol.~24, pp.
  342--357, 2019.

\bibitem{vidhi2018review}
R.~Vidhi and P.~Shrivastava, ``A review of electric vehicle lifecycle emissions
  and policy recommendations to increase ev penetration in india,''
  \emph{Energies}, vol.~11, no.~3, p. 483, 2018.

\bibitem{gupta2020sustainable}
D.~Gupta and A.~Garg, ``Sustainable development and carbon neutrality:
  Integrated assessment of transport transitions in india,''
  \emph{Transportation Research Part D: Transport and Environment}, vol.~85, p.
  102474, 2020.

\bibitem{sarangi2018green}
G.~K. Sarangi, ``Green energy finance in india: Challenges and solutions,''
  ADBI Working Paper Series, Tech. Rep., 2018.

\bibitem{CHEN2020}
``The influence of increased population density in china on air pollution,''
  \emph{Science of The Total Environment}, vol. 735, p. 139456, 2020.

\bibitem{ZHANG2020}
Z.~Zhang, T.~Xue, and X.~Jin, ``Effects of meteorological conditions and air
  pollution on covid-19 transmission: Evidence from 219 chinese cities,''
  \emph{Science of The Total Environment}, vol. 741, p. 140244, 2020.

\bibitem{chandra2019langevin}
R.~Chandra, K.~Jain, R.~V. Deo, and S.~Cripps, ``Langevin-gradient parallel
  tempering for {Bayesian} neural learning,'' \emph{Neurocomputing}, vol. 359,
  pp. 315--326, 2019.

\bibitem{chandra2020bayesian}
R.~Chandra and A.~Kapoor, ``Bayesian neural multi-source transfer learning,''
  \emph{Neurocomputing}, vol. 378, pp. 54--64, 2020.

\bibitem{chandra2017comtltime}
R.~Chandra, Y.-S. Ong, and C.-K. Goh, ``Co-evolutionary multi-task learning
  with predictive recurrence for multi-step chaotic time series prediction,''
  \emph{Neurocomputing}, vol. 243, pp. 21--34, 2017.

\bibitem{chandra2021deepcovid}
R.~Chandra, A.~Jain, and D.~S. Chauhan, ``Deep learning via {LSTM} models for
  {COVID-19} infection forecasting in india,'' \emph{arXiv preprint
  arXiv:2101.11881}, 2021.

\end{thebibliography}







\end{document}